%% file: neurips_2026.tex
\documentclass{article}

\PassOptionsToPackage{numbers, sort}{natbib}
\usepackage[preprint]{neurips_2026}

\usepackage[utf8]{inputenc} % allow utf-8 input
\usepackage[T1]{fontenc}    % use 8-bit T1 fonts
\usepackage{hyperref}       % hyperlinks
\usepackage{url}            % simple URL typesetting
\usepackage{booktabs}       % professional-quality tables
\usepackage{amsfonts}       % blackboard math symbols
\usepackage{nicefrac}       % compact symbols for 1/2, etc.
\usepackage{microtype}      % microtypography
\usepackage[dvipsnames]{xcolor}         % colors
\usepackage{enumitem}
\usepackage{graphicx}
\usepackage{amssymb}
\usepackage{booktabs}
\usepackage{tabularx}
\usepackage{multirow}
\usepackage{wrapfig}
\usepackage{xspace}
\usepackage{array}
\usepackage{subcaption}

\usepackage[most]{tcolorbox}
\newcommand{\Thud}{\textsc{Thud}\xspace}

\definecolor{tradeofffg}{HTML}{2F6F8F}
\definecolor{oursfg}{HTML}{8C2D1F}
\definecolor{syncbg}{HTML}{EDF4FF}
\definecolor{syncfg}{HTML}{3568A8}

\definecolor{videobg}{HTML}{EEF7F0}
\definecolor{videofg}{HTML}{3E7C4A}

\definecolor{omnibg}{HTML}{FCF6EF}
\definecolor{omnifg}{HTML}{AF6A34}

\definecolor{avgbg}{HTML}{F6F1E3}
\definecolor{avgfg}{HTML}{8A6A2B}

\newcolumntype{S}{>{\columncolor{syncbg}}c}
\newcolumntype{V}{>{\columncolor{videobg}}c}
\newcolumntype{O}{>{\columncolor{omnibg}}c}
\newcolumntype{A}{>{\columncolor{avgbg}}c}

\definecolor{verifyfg}{HTML}{2A7F9E}
\definecolor{difficulty}{HTML}{6A3D9A}
\definecolor{pipelineblue}{HTML}{1F4E8C}
\definecolor{alignorange}{HTML}{B45309}
\definecolor{failurered}{HTML}{991B1B}

\definecolor{heatmap}{HTML}{8B1E1E}
\definecolor{breakdownblue}{HTML}{2A5C99}

\usepackage[table]{xcolor}
\definecolor{shiftbg}{HTML}{EAF1FF}
\definecolor{mutebg}{HTML}{F0EAFF}
\definecolor{swapbg}{HTML}{EAF7EE}
\definecolor{gapbg}{HTML}{FFF1CC}

\tcbset{
  gptbox/.style={
    colback=gray!3, colframe=gray!55, coltitle=black,
    colbacktitle=gray!55,
    fonttitle=\bfseries\small, boxrule=0.5pt, arc=2pt,
    left=5pt, right=5pt, top=4pt, bottom=4pt,
    breakable, enhanced,
  },
  videoline/.style={
    colback=gray!8, colframe=gray!30, boxrule=0pt, arc=1pt,
    left=4pt, right=4pt, top=2pt, bottom=2pt,
    fontupper=\footnotesize\ttfamily,
  },
}

\tcbset{
  gptbox/.style={
    colback=gray!3, colframe=gray!55, coltitle=white,
    colbacktitle=gray!55,
    fonttitle=\bfseries\small, boxrule=0.5pt, arc=2pt,
    left=5pt, right=5pt, top=4pt, bottom=4pt,
    breakable, enhanced,
  },
  videoline/.style={
    colback=gray!8, colframe=gray!30, boxrule=0pt, arc=1pt,
    left=4pt, right=4pt, top=2pt, bottom=2pt,
    fontupper=\footnotesize\ttfamily,
  },
  promptbox/.style={
    colback=blue!2, colframe=blue!40!black, coltitle=white,
    colbacktitle=blue!40!black,
    fonttitle=\bfseries\small, boxrule=0.5pt, arc=2pt,
    left=5pt, right=5pt, top=4pt, bottom=4pt,
    breakable, enhanced,
    fontupper=\small,
  },
  judgebox/.style={
    colback=orange!3, colframe=orange!60!black, coltitle=white,
    colbacktitle=orange!60!black,
    fonttitle=\bfseries\small, boxrule=0.5pt, arc=2pt,
    left=5pt, right=5pt, top=4pt, bottom=4pt,
    breakable, enhanced,
    fontupper=\small,
  },
}

\definecolor{shiftblue}{HTML}{2F6DAE}
\definecolor{mutepurple}{HTML}{6A3D9A}
\definecolor{swapgreen}{HTML}{3F7D3D}

\definecolor{mazecol}{HTML}{F2F1E7}
\definecolor{flowcol}{HTML}{37b5ac}
\definecolor{sokocol}{HTML}{81B29A}

\colorlet{temporalcol}{mazecol}
\colorlet{audiocol}{flowcol!25}
\colorlet{soundcol}{sokocol!50}

\newcommand{\shift}{\textcolor{shiftblue}{\textbf{\textit{Shift}}}\xspace}
\newcommand{\mute}{\textcolor{mutepurple}{\textbf{\textit{Mute}}}\xspace}
\newcommand{\swap}{\textcolor{swapgreen}{\textbf{\textit{Swap}}}\xspace}

\definecolor{darkblue}{rgb}{0, 0, 0.5}
\definecolor{bestcolor}{rgb}{0.0, 0.45, 0.0}
\definecolor{ourpurple}{RGB}{230,220,245}
\definecolor{citecolor}{HTML}{0071BC}
\definecolor{linkcolor}{HTML}{CC5A57}
\tcbuselibrary{skins}
\newtcolorbox{promptbox}[1]{breakable,enhanced,colback=gray!5,colframe=gray!55,fonttitle=\bfseries,title=#1,left=4pt,right=4pt,top=4pt,bottom=4pt}
\hypersetup{colorlinks=true, citecolor=citecolor, linkcolor=linkcolor, urlcolor=darkblue}

\definecolor{descBlue}{HTML}{E7EFFA}
\definecolor{descBlueLine}{HTML}{82B0D2}

\definecolor{locGreen}{HTML}{E8F5EF}
\definecolor{locGreenLine}{HTML}{8ECFC9}

\definecolor{attrPurple}{HTML}{F0EEF8}
\definecolor{attrPurpleLine}{HTML}{BEB8DC}

\definecolor{qaOrange}{HTML}{FFF2E2}
\definecolor{qaOrangeLine}{HTML}{FFBE7A}

\definecolor{bluegray}{HTML}{64748B}

\definecolor{ucdcolor}{HTML}{1F4E8C}
\definecolor{princetoncolor}{HTML}{E77500}
\definecolor{wisconsincolor}{HTML}{C5050C}
\definecolor{uniphorecolor}{HTML}{2E8B57}

\newcommand{\ucd}{{\color{ucdcolor}\boldsymbol{d}}}
\newcommand{\princeton}{{\color{princetoncolor}\boldsymbol{p}}}
\newcommand{\wisconsin}{{\color{wisconsincolor}\boldsymbol{w}}}
\newcommand{\uniphore}{{\color{uniphorecolor}\boldsymbol{u}}}

\title{When Vision Speaks for Sound}
\author{
Xiaofei Wen$\hspace{.1em}^{\ucd}$ \hspace{.8em}
Wenjie Jacky Mo$\hspace{.1em}^{\ucd}$ \hspace{.8em}
Xingyu Fu$\hspace{.1em}^{\princeton}$ \hspace{.8em}
Rui Cai$\hspace{.1em}^{\ucd}$
\vspace{.2em} \\
\textbf{Tinghui Zhu}$\hspace{.1em}^{\ucd}$ \hspace{.8em}
\textbf{Wendi Li}$\hspace{.1em}^{\wisconsin}$ \hspace{.8em}
\textbf{Yanan Xie}$\hspace{.1em}^{\uniphore}$ \hspace{.8em}
\textbf{Muhao Chen}$\hspace{.1em}^{\ucd}$ \hspace{.8em}
\textbf{Peng Qi}$\hspace{.1em}^{\uniphore}$
\vspace{.5em} \\
$\hspace{.1em}^{\ucd}$University of California, Davis
\hspace{1.5em}
$\hspace{.1em}^{\princeton}$Princeton University
\vspace{.2em} \\
$\hspace{.1em}^{\wisconsin}$University of Wisconsin--Madison
\hspace{1.5em}
$\hspace{.1em}^{\uniphore}$Uniphore \\
\vspace{0.1em} \\ 
    \texttt{Website:} \href{https://rakanwen.github.io/when-vision-speaks-for-sound/}{\texttt{when-vision-speaks-for-sound}}
    \hspace{1em}
    \raisebox{-0.4ex}{\includegraphics[height=1em]{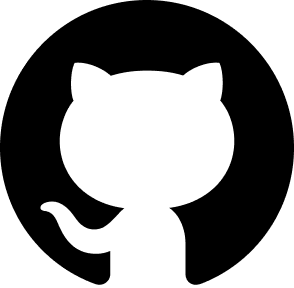}}\hspace{0.3em}\href{https://github.com/rakanWen/wvs-code}{\texttt{Code}} 
    \hspace{0.2cm}
    \raisebox{-0.4ex}{\includegraphics[height=1em]{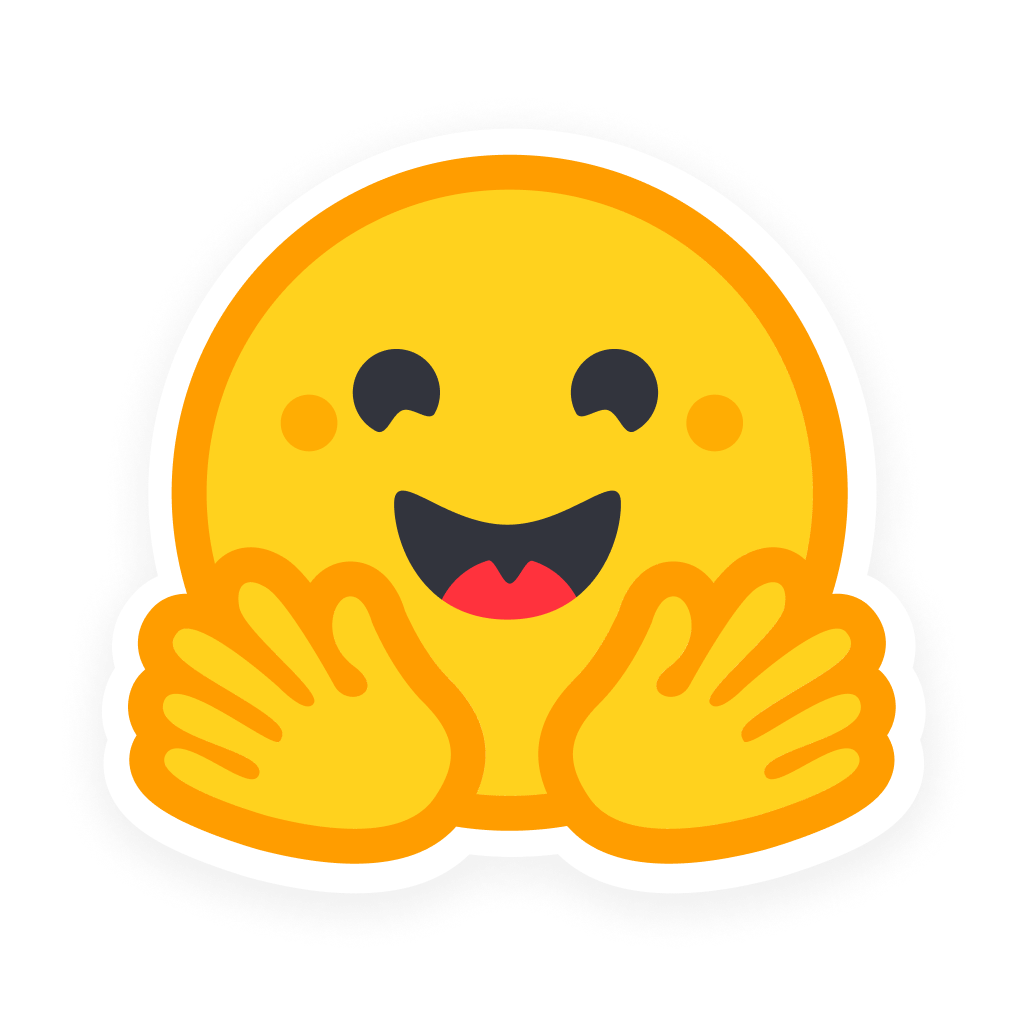}}\hspace{0.3em}\href{https://huggingface.co/Rakancorle1/wvs-thud-model}{\texttt{Model}}
}

\usepackage{cleveref}
\Crefformat{figure}{#2Fig.~#1#3}
\Crefmultiformat{figure}{Figs.~#2#1#3}{ and~#2#1#3}{, #2#1#3}{ and~#2#1#3}
\Crefformat{table}{#2Tab.~#1#3}
\Crefmultiformat{table}{Tabs.~#2#1#3}{ and~#2#1#3}{, #2#1#3}{ and~#2#1#3}
\Crefformat{appendix}{Appx.~\S#2#1#3}
\Crefformat{section}{\S#2#1#3}
\Crefformat{subsection}{\S#2#1#3}
\crefformat{algorithm}{Alg.~#2#1#3}
\crefformat{equation}{Eq.~#2#1#3}
\crefformat{enumi}{Prop.~#2#1#3}
\crefformat{listing}{Code.~#2#1#3}
\crefformat{lstlisting}{Code.~#2#1#3}
\Crefmultiformat{enumi}{Prop.~#2#1#3}{ and~#2#1#3}{, #2#1#3}{ and~#2#1#3}

\begin{document}

\maketitle

\input{sections/00_Abstract}

\input{sections/01_Introduction}

\input{sections/03_InterAlign}

\input{sections/04_Experiment}

\input{sections/02_Related_Work}
\input{sections/05_Conclusion}
\newpage
{
\small
\bibliographystyle{plain}
\bibliography{sections/nips2026_conference}
}
\clearpage
\appendix

\input{sections/Appendix}

\clearpage

\end{document}

%% file: sections/00_Abstract.tex
\vspace{-1.5em}
\begin{abstract}

\looseness-1
Despite rapid progress in video-capable MLLMs, we find that their apparent audio understanding in videos is often vision-driven: models rely on visual cues to infer or hallucinate acoustic information, rather than verifying the audio stream. This issue appears across both state-of-the-art open-source omni models and leading closed-source models from providers such as Google and OpenAI. We characterize this failure mode as an audio-visual \textit{Clever Hans} effect, in which models appear (falsely) audio-grounded, but actually exploit visual-acoustic correlations without verifying whether the audio and visual streams are truly aligned.
To systematically study this behavior, we introduce \Thud, an intervention-driven probing framework based on three counterfactual audio edits: \textit{Shift}, which tests temporal synchronization; \textit{Mute}, which tests sound existence; and \textit{Swap}, which tests audio-visual consistency. Beyond diagnosis, we further study a two-stage alignment recipe: intervention-derived preference pairs teach audio verification, while event-level general video preferences regularize the model against over-specialization. Our best 10K-sample recipe improves average performance across the three intervention dimensions by 28 percentage points, while slightly improving performance on general video and audio-visual QA benchmarks.

% \peng{the abstract should briefly mention that our findings apply to both SOTA open-source models and leading closed source models from providers like Google, OpenAI, and Anthropic}
\end{abstract}

%% file: sections/01_Introduction.tex
\section{Introduction}
\vspace{-0.5em}
% The Illusion of Progress
Multimodal Large Language Models (MLLMs) have rapidly advanced video understanding~\cite{DBLP:conf/emnlp/LinYZCNJ024, DBLP:conf/acl/0001RKK24, DBLP:journals/tmlr/ZhangWLLMLL25}. Powered by foundation models such as GPT~\cite{DBLP:journals/corr/abs-2601-03267}, Gemini~\cite{deepmind2026gemini3}, and Qwen-VL~\cite{DBLP:journals/corr/abs-2511-21631}, recent Video-LLMs~\cite{DBLP:conf/nips/Dai0LTZW0FH23, zhang-etal-2023-video, Li2023VideoChatCV, DBLP:journals/corr/abs-2504-10479} can interpret dynamic scenes~\cite{Fu2024BLINKML, Ren2024Location}, answer complex questions~\cite{patraucean2023perception,Li2024MVBench}, and follow instructions~\cite{Wang2024InternVideo2SV,jin2023chatunivi}. 
Yet, in videos with both visual and acoustic signals, such capabilities can blur the boundary between genuine audio-visual grounding and visually driven narration. % such capability can conflate audio-visual grounding with visual narration: 
For example, when shown a skateboarder crashing onto concrete, a model may describe a heavy thud even when the audio evidence is absent or misaligned~\cite{DBLP:conf/emnlp/LiDZWZW23,DBLP:conf/cvpr/GuanLWXLL0CHYM024,sung-bin2025avhbench,Cai2025DiagnosingAM}.
Such behavior is often interpreted as multimodal perception, but it may instead reflect an illusion of audio-visual understanding: the model predicts what a video should sound like from what it sees. 
While static vision-language models are known to behave like ``bags-of-words'' driven by text priors~\cite{DBLP:conf/cvpr/Tong0Z0LX24, DBLP:conf/cvpr/ThrushJBSWKR22, DBLP:conf/iclr/Yuksekgonul0KJ023}, analogous prediction shortcuts in dynamic audio-visual contexts remain underexplored. This raises a central question: \textit{Are current video-capable multimodal models truly performing audio-visual grounding, or merely hallucinating acoustic events from visual-semantic shortcuts?}

\begin{tcolorbox}[
colback=bluegray!4,
colframe=bluegray!35,
boxrule=0.5pt,
arc=2pt,
left=6pt,
right=6pt,
top=3pt,
bottom=3pt,
title=\textbf{Clever Hans Effect in Audio-Visual Grounding},
fonttitle=\small,
coltitle=black
]
\looseness-3
\small
A video-capable multimodal model exhibits a \textit{Clever Hans effect} when it appears audio-grounded but produces sound-related outputs primarily from visual cues rather than verified audio evidence.
\end{tcolorbox}

\begin{figure}[t]
    \centering
    \includegraphics[width=\linewidth]{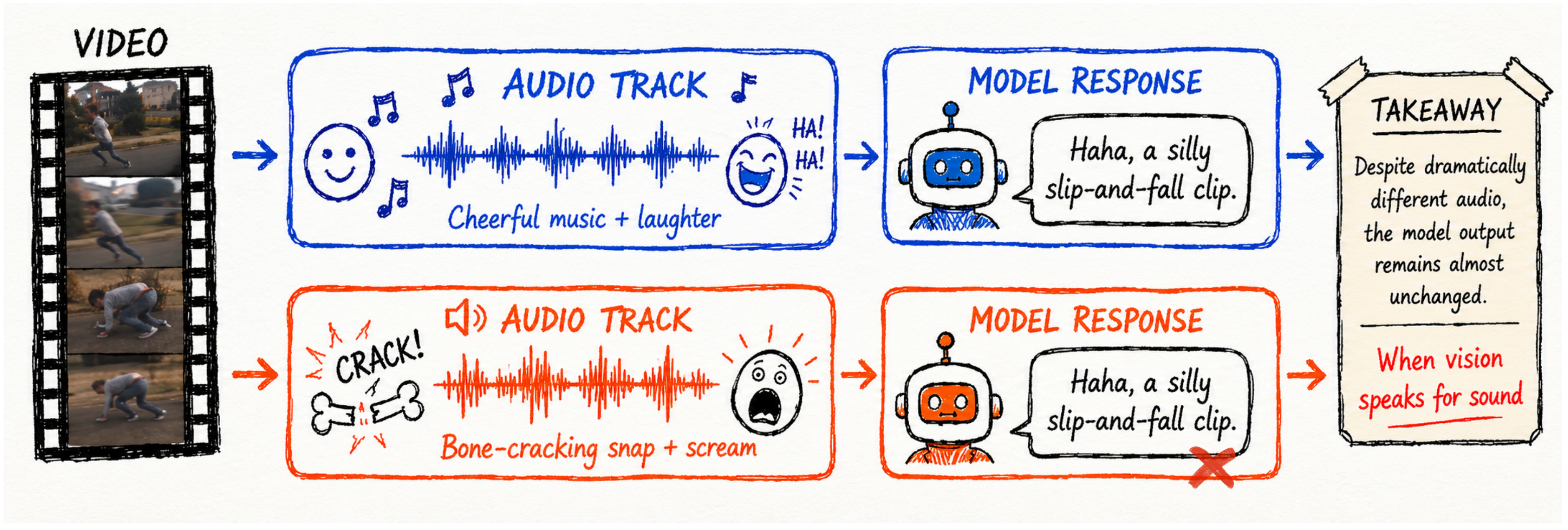}

    \caption{
\textcolor{failurered}{\textbf{When vision speaks for sound}}.
Given the same visual event but different audio tracks, current video-capable models produce nearly identical captions, suggesting visual-prior shortcutting rather than audio-grounded understanding.
}
\vspace{-1cm}

    \label{fig:motivation}

\end{figure}

% \vspace{-0.3em}

\looseness-1
We find that current video-capable MLLMs are often visually dominated when reasoning about audio-related information in sounded videos. As illustrated in \Cref{fig:motivation}, this shortcut can lead models to produce nearly unchanged descriptions even when the audio track changes substantially. This behavior resembles the famous \textit{Clever Hans effect}~\cite{pfungst1911clever}, where apparent competence arises from exploiting unintended but correlated cues rather than performing the intended task. Such semantic laziness~\cite{DBLP:journals/natmi/GeirhosJMZBBW20} allows models to exploit visual-semantic shortcuts and language priors instead of fine-grained audio-visual grounding that checks whether the audio and visual streams are temporally and semantically consistent~\cite{DBLP:conf/iclr/Yuksekgonul0KJ023, DBLP:conf/cvpr/GoyalKSBP17}. This failure often remains hidden because common audio-visual evaluations preserve the natural correlations that make such shortcuts effective~\cite{Gemmeke2017AudioSA,Chen2020VggsoundAL,Carreira2017QuoVA}: barking dogs produce barks, falling objects produce impacts, and speaking faces produce speech~\cite{Arandjelovi2017LookLA,Owens2018AudioVisualSA}. As a result, a model can appear grounded by recognizing the visual event and predicting its likely sound, without verifying whether that sound is actually present, synchronized, or physically consistent. This pseudo-alignment creates an illusion of multimodal understanding that current evaluations often fail to expose~\cite{DBLP:conf/nips/MangalamAM23,DBLP:conf/cvpr/0002WH00LWX0L0024}. To expose the Clever Hans effect, evaluation must move beyond naturally correlated videos and use controlled interventions that systematically break the audio-visual correspondences that allow visual-semantic shortcuts to succeed~\cite{Korbar2018CooperativeLO,Morgado2020AudioVisualID}.

To this end, we introduce \textbf{\Thud}~(\textbf{T}emporal and \textbf{H}allucination \textbf{U}nmasking \textbf{D}iagnostics), an intervention-driven diagnostic protocol for probing audio-visual grounding in sounded videos. \Thud~constructs a dynamic probing space by counterfactually perturbing the audio-visual correspondences of natural videos across temporal synchronization, audio existence, and sound consistency, thereby neutralizing semantic shortcuts and exposing whether a model engages in genuinely grounded audio-visual reasoning or merely hallucinates from visual-semantic and language priors. Beyond diagnosis, we further study whether targeted post-training can mitigate these shortcuts through a family of alignment recipes that combine intervention-derived preference pairs with general video data. The best-performing recipe uses a 10K-sample mixture of counterfactual temporal preferences and event-level general video supervision, substantially improving the model's ability to detect temporal interventions, including out-of-distribution synchronization tests, while avoiding an alignment tax~\cite{Askell2021AGL,DBLP:conf/nips/Ouyang0JAWMZASR22} on standard video understanding benchmarks. 
Additional targeted supervision on Mute and Swap further improves audio-existence and sound-consistency verification, showing that intervention-based training can be extended beyond temporal alignment.
However, the same training yields only marginal gains without such targeted examples, suggesting that temporal synchronization, audio existence, and sound consistency are distinct failure modes of grounded audio-visual understanding rather than a single unified deficiency.

In summary, we make three contributions:
1) We identify and systematically expose a \textit{Clever Hans} effect in current Video-LLMs, where models substitute genuine audio-visual grounding with visual-semantic shortcuts. Through controlled interventions, we quantify how strongly models rely on visual priors when answering sound-related questions.
2) We introduce \Thud, a counterfactual diagnostic protocol that dismantles natural cross-modal correlations. By applying \textit{Mute}, \textit{Shift}, and \textit{Swap} interventions, \Thud{} audits existential, temporal, and material aspects of audio-visual grounding.
3) We evaluate preference-optimization recipes for mitigating audio-visual shortcuts. Our final 10K recipe improves average performance across \textit{Shift}, \textit{Mute}, and \textit{Swap} interventions by 28\%, while slightly improving general video and audio-visual understanding.

% Contributions
% In summary, our contributions are three-fold:
% \begin{itemize}[leftmargin=1.5em, itemsep=4pt, topsep=4pt, parsep=0pt]
%     \item We identify and systematically expose a \textit{Clever Hans} effect in current Video-LLMs, where models can substitute genuine audio-visual grounding with visual-semantic shortcuts. Through controlled interventions, we quantify how strongly these models rely on visual priors when answering sound-related questions.
    
%     \item We introduce \Thud, a counterfactual diagnostic protocol that systematically dismantles natural cross-modal correlations. By applying \textit{Mute}, \textit{Shift}, and \textit{Swap} interventions, \Thud{} forces models into counterfactual settings that explicitly audit existential, temporal, and material aspects of audio-visual grounding.
    
%     \item We evaluate a family of preference-optimization recipes for mitigating audio-visual shortcuts. Our final 10K recipe improves average performance across \textit{Shift}, \textit{Mute}, and \textit{Swap} interventions by 28\%, while slightly improving general video understanding and audio-visual understanding.
% \end{itemize}

%% file: sections/03_InterAlign.tex
% \section{Intervention-Driven Alignment for Audio-Visual Grounding}
\section{How Can We Align Models Beyond Visual Shortcuts?}
\label{sec:method}

\begin{figure}[t]
    \centering
    \includegraphics[width=\linewidth]{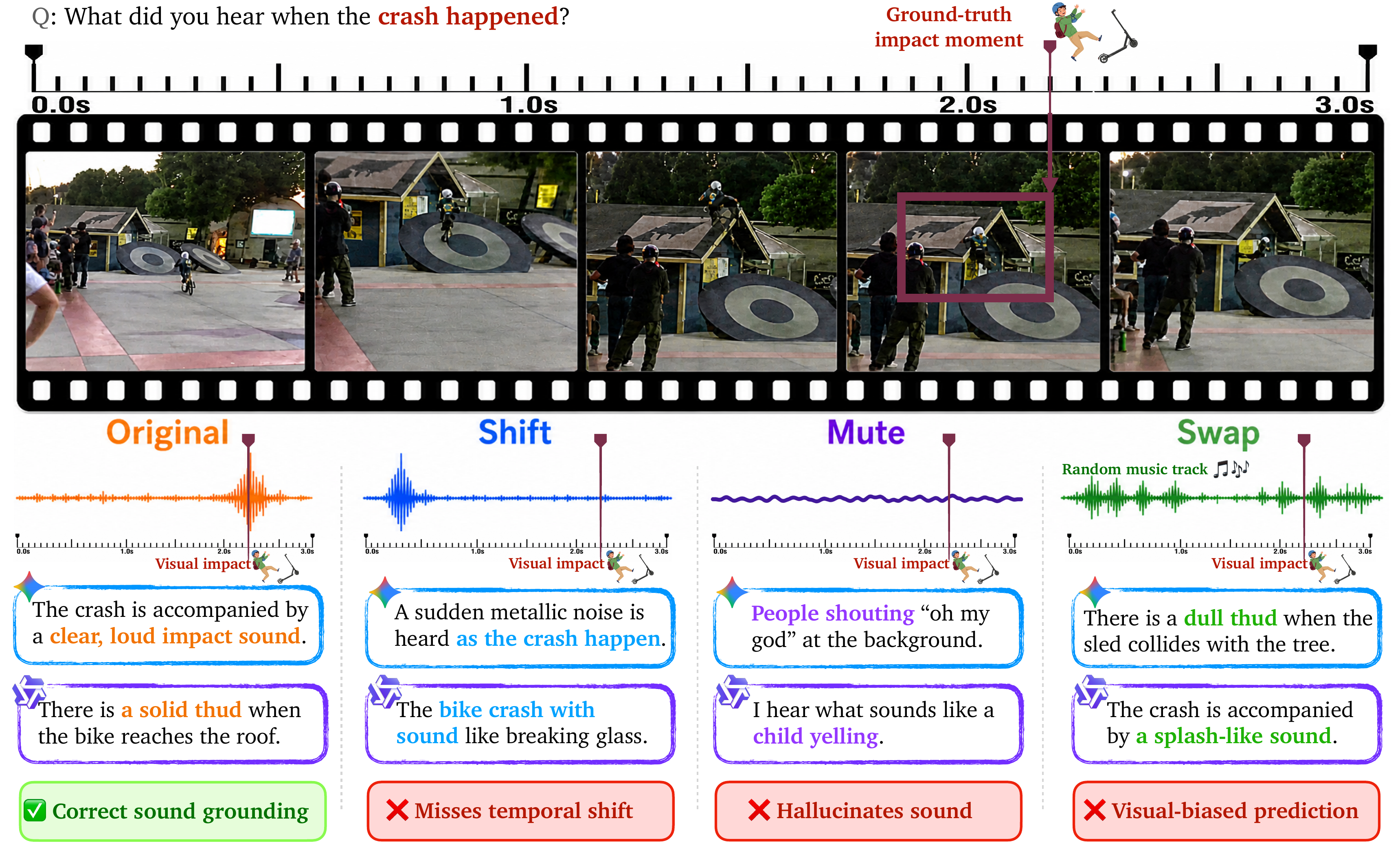}
    \vspace{-1em}
    \caption{
    \textcolor{failurered}{\textbf{Representative failure cases}} under \textit{Shift}, \textit{Mute}, and \textit{Swap} interventions.
    Gemini and Qwen3-Omni often rely on visual priors rather than verifying the audio stream, leading to missed temporal shifts, hallucinated sounds, and visually biased predictions.
    }
    \label{fig:pilot_cases}
    \vspace{-1em}
\end{figure}

\Cref{fig:pilot_cases} illustrates that even native multimodal models such as Gemini and Qwen3-Omni can produce plausible acoustic interpretation from visual actions alone, rather than verifying %audio presence, timing, or source consistency. 
whether the corresponding sound is present, temporally aligned, or consistent with its visual source.
These failures motivate our intervention-driven diagnostic protocol, which deliberately breaks natural audio-visual correlations to expose models' reliance on visual-semantic shortcuts.

To align models beyond visual shortcuts, we construct training signals that task them to compare visible events against the actual audio stream rather than rely on visual priors. Our recipe turns physical audio-visual interventions into alignment data in three steps. First, we source videos with salient acoustic consequences and break natural correlations (\Cref{sec:dspi}). Second, we annotate event-time labels and construct chosen--rejected preference pairs (\Cref{sec:AAPC}). And third, we combine intervention data with general video instruction data to preserve overall comprehension (\Cref{sec:MGVID}).

\subsection{Data Sourcing and Physical Interventions}
\label{sec:dspi}

To build intervention data for audio-visual grounding, we use the Oops dataset~\cite{Epstein2019OopsPU}, a collection of in-the-wild videos centered on unintentional human actions. As shown in \Cref{app:data-construction-pipeline}, Oops contains many failure-centered events, such as slipping, skiing crashes, and objects breaking, that naturally induce strong expectations about the accompanying sound. This property makes it a suitable source for constructing Clever Hans-style cases: the visual content often suggests a plausible acoustic event, while the audio track determines whether that event is actually present, temporally aligned, and physically consistent with the observed action.

\paragraph{Formalizing interventions.}
Let a video be represented as $v=(x_{1:T}, a_{1:T})$, where $x_{1:T}$ denotes the visual stream and $a_{1:T}$ denotes the audio track. We construct intervened videos by applying one of three operators:
\begin{equation}
\tilde{v} =
\mathcal{I}_{k}(v), \quad
k \in \{\shift, \mute, \swap\}.
\end{equation}
For \shift, the audio track is displaced by a temporal offset $\Delta$:
\begin{equation}
\mathcal{I}_{\textsc{Shift}}(v;\Delta) = (x_{1:T}, a_{1:T}^{+\Delta}),
\quad \Delta \in [-\Delta_{\max}, \Delta_{\max}].
\end{equation}
Here, $\Delta<0$ corresponds to an \textit{early} audio event, while $\Delta>0$ corresponds to a \textit{delayed} audio event. This intervention requires the model to compare the timing of the visible event with the timing of its acoustic consequence.

For \mute, the audio signal is replaced with silence:
\begin{equation}
\mathcal{I}_{\textsc{Mute}}(v) = (x_{1:T}, \varnothing).
\end{equation}
For \swap, the original audio is replaced with an audio track $a'_{1:T}$ from another video:
\begin{equation}
    \mathcal{I}_{\textsc{Swap}}(v, v')
    =
    (x_{1:T}, a'_{1:T}),
    \qquad
    v' = (x'_{1:T}, a'_{1:T}).
\end{equation}
The substituted audio is acoustically plausible but physically inconsistent with the visible event, forcing the model to verify audio-visual consistency rather than rely on the most likely sound implied by vision alone.
Overall, these interventions convert naturally correlated videos into controlled counterfactual cases that target temporal synchronization, sound presence, and physical consistency; a detailed summary is provided in \Cref{app:intervention_summary}.

% \begin{table}[t]
% \caption{
% Summary of our three physical interventions. Each intervention breaks a different natural audio-visual correlation and poses a diagnostic question about audio-visual grounding.
% }
% \label{tab:intervention-summary}

% \centering
% \begin{tabular}{lccc}
% \toprule
% \textbf{Intervention} 
% & \textbf{Operation} 
% & \textbf{Broken correlation}
% & \textbf{Diagnostic question} \\
% \midrule
% \shift
% & $a_{1:T} \rightarrow a_{1:T}^{+\Delta}$ 
% & temporal synchrony
% & Is the sound synchronized? \\
% \mute 
% & $a_{1:T} \rightarrow \varnothing$ 
% & sound existence
% & Is any sound present? \\
% \swap
% & $a_{1:T} \rightarrow a'_{1:T}$ 
% & source consistency
% & Does the sound match the event? \\
% \bottomrule
% \end{tabular}

% \vspace{-0.5em}
% \end{table}

\subsection{Annotation and Preference Pair Construction}
\label{sec:AAPC}

We annotate each source video with event-time labels used to evaluate audio-visual interventions:
\begin{equation}
    z_i = (e_i^v, t_i^v, e_i^a, t_i^a),
\end{equation}
where $e_i^v$ and $t_i^v$ denote the visual event and its timestamp, $e_i^a$ and $t_i^a$ denote the corresponding acoustic event and timestamp. These fields correspond to the visual event, visual time, audio event, and audio time labels in \Cref{fig:data-construction} (\Cref{app:data-construction-pipeline}).

\paragraph{Cross-model verification.}
We use Gemini to generate initial event-time annotations because it supports direct video ingestion and can inspect both visual and audio streams. For visual timestamps, we further verify Gemini's annotations with GPT and Claude by decomposing each video into $N$ temporally ordered frame units and asking the models to locate the visual event within the frame sequence. For audio timestamps, which require access to the acoustic stream, we cross-verify Gemini's predictions with human inspection.

Let $\mathcal{M}_v$ denote the set of visual annotator models and let $\mathcal{M}_a=\{\mathrm{Gemini}, \mathrm{Human}\}$ denote the audio verification sources.
\begin{equation}
    z_i^{(m)} = 
    \left(
    e_i^{v,m}, t_i^{v,m},
    e_i^{a,m}, t_i^{a,m}
    \right),
\end{equation}
where visual fields are available for $m \in \mathcal{M}_v$ and audio fields are available for $m \in \mathcal{M}_a$. A sample is automatically retained when both visual and acoustic timestamps agree within strict tolerances:
\begin{equation}
    \max_{m,m' \in \mathcal{M}_v}
    \left|t_i^{v,m} - t_i^{v,m'}\right|
    \leq \epsilon_v,
    \qquad
    \max_{m,m' \in \mathcal{M}_a}
    \left|t_i^{a,m} - t_i^{a,m'}\right|
    \leq \epsilon_a.
\end{equation}
Here, $\epsilon_v$ and $\epsilon_a$ denote the tolerance thresholds for visual and acoustic timestamps, respectively. Cases with model disagreement are manually inspected and corrected to ensure reliable event-time labels. We provide the annotation prompts, frame-unit construction details, agreement criteria, and manual verification protocol in \Cref{app:annotation}.

\paragraph{Preference pair construction.}
The annotated intervention cases are converted into chosen--rejected preference pairs:
\begin{equation}
    \mathcal{D}_{\mathrm{pref}}
    =
    \left\{
    \left(
    \tilde{v}_i, q_i, y_i^+, y_i^-
    \right)
    \right\}_{i=1}^{N},
\end{equation}
where $\tilde{v}_i$ is the intervened video, $q_i$ is the diagnostic prompt, $y_i^+$ is the chosen response, and $y_i^-$ is the rejected response. The chosen response explicitly verifies the audio-visual relation, while the rejected response is visually plausible but inconsistent with the audio evidence, approximating the shortcut behavior we aim to suppress. The overall annotation and intervention pipeline is summarized in \Cref{fig:data-construction} (\Cref{app:data-construction-pipeline}).

For \shift, chosen responses detect early or delayed audio, while rejected responses claim synchronization or the wrong temporal direction. For \mute, chosen responses identify silence, while rejected responses hallucinate expected sounds. For \swap, chosen responses flag audio-visual source inconsistency, while rejected responses accept the mismatched sound. These pairs train the model to verify audio evidence rather than follow visually plausible shortcuts. Examples are provided in \Cref{app:preference_examples}.

\subsection{Two-Stage Alignment with General Video Data}
\label{sec:MGVID}

Intervention data provides targeted supervision for detecting \shift, \mute, and \swap failures, but may over-specialize the model to counterfactual cases. We therefore mix it with general video instruction data, whose temporally segmented annotations expose ordinary audio-visual correspondences at the event level. \Cref{app:alignment-pipeline} summarizes this two-stage alignment pipeline.

We use FineVideo~\cite{Farre2024FineVideo} as the source of general video data because its annotations are organized around time segments, describing what occurs from one timestamp range to the next. We re-annotate selected FineVideo clips with Gemini and apply human agreement checks, enriching the original segment annotations with both visual and audible event-level information. The resulting annotations are used to construct four instruction types summarized in \Cref{app:finevideo-data}.

% \begin{tcolorbox}[
% colback=gray!3,
% colframe=gray!35,
% boxrule=0.6pt,
% arc=2pt,
% left=5pt,
% right=5pt,
% top=4pt,
% bottom=4pt
% ]
% \small
% \textbf{FineVideo-derived instruction data.}
% We construct four complementary data types:
% \textit{description} for visible and audible content,
% \textit{localization} for event timestamps,
% \textit{attribution} for source or material cues,
% and \textit{audio-dependent QA} for questions requiring audio evidence.
% \end{tcolorbox}

Our training follows the standard post-training recipe of Supervised Fine Tuning (SFT) followed by preference alignment~\cite{DBLP:conf/nips/ChristianoLBMLA17, DBLP:journals/corr/abs-1909-08593, DBLP:conf/nips/Ouyang0JAWMZASR22}. 
We use SFT warm-up on intervention-derived data to establish audio-aware response patterns, and then apply DPO on intervention preference pairs mixed with general video data to favor audio-verified responses over visually plausible shortcuts. The general video mixture is included to reduce over-specialization to intervention cases and preserve broad video understanding. The overall two-stage alignment pipeline is summarized in \Cref{fig:preference_optimization} (\Cref{app:alignment-pipeline}).

%% file: sections/04_Experiment.tex
\vspace{-4pt}
\section{Experiments}
\label{sec:exp}

This section presents the experiments for diagnosing audio-visual shortcut reliance and evaluating targeted alignment, covering the setup (\Cref{ssec:setup}), shortcut analysis (\Cref{sec:shortcut_diagnosis}), targeted alignment improvements (\Cref{sec:alignment_tax}), and broader intervention results (\Cref{sec:beyond_temporal}).

\subsection{Experimental Setup}\label{ssec:setup}

\paragraph{Evaluation conditions and metrics.}
We evaluate audio-visual grounding under four conditions: \textit{Original}, \shift, \mute and \swap. Original videos serve as positive controls with natural audio-visual correspondence, while the interventions probe audio existence, temporal synchronization, and sound consistency. We report paired accuracy for each grounding dimension.

\looseness-2

\paragraph{Models.}
We group evaluated models by access mode. The API-tested models include Gemini-3.1-Pro~\cite{deepmind2026gemini3}, MiMo-V2.5~\cite{xiaomi2026mimov25}, and Nemotron-3-Nano-Omni~\cite{Deshmukh2026Nemotron3N}. We also query GPT-5.5~\cite{DBLP:journals/corr/abs-2601-03267}, but omit it from~\Cref{tab:shortcut_diagnosis} because its tested interface does not support direct audio input for video; its outputs are provided in \Cref{app:gpt-qualitative}. The locally evaluated models include MiniCPM-o-4.5~\cite{Cui2026MiniCPMo4T}, Qwen3-Omni~\cite{DBLP:journals/corr/abs-2509-17765}, and Ming-flash-omni-2.0~\cite{DBLP:journals/corr/abs-2506-09344}.

\looseness-2

\paragraph{Training and general capability evaluation.}
For controlled training experiments, we use Qwen3-Omni-30B as the trainable backbone and compare checkpoints trained with different combinations of intervention data and general video data. To test whether intervention training incurs an alignment tax, we evaluate these checkpoints on Video-MME~\cite{DBLP:conf/cvpr/FuDLLRZWZSZCLLZ25}, LVBench~\cite{DBLP:journals/corr/abs-2406-08035}, DailyOmni~\cite{DBLP:journals/corr/abs-2505-17862}, and WorldSense~\cite{DBLP:journals/corr/abs-2502-04326}, which measure general video and omni-modal understanding beyond our intervention distribution. We further evaluate on VGGSoundSync~\cite{Chen21b} to test out-of-distribution temporal synchronization beyond our constructed intervention set.

\begin{table}[t]
\caption{
Paired diagnostic accuracy (\%) of video-capable multimodal models.
\textbf{Orig.} denotes naturally correlated controls, while \shift, \mute, and \swap denote counterfactual interventions.
\textbf{Avg Gap} is the average accuracy drop, reflecting shortcut reliance.
}
\label{tab:shortcut_diagnosis}
\centering
\small
\begin{tabular}{lcccccccc}
\toprule
\multirow{2}{*}{\textbf{Model}}
& \multirow{2}{*}{\textbf{Size}}
& \multicolumn{2}{c}{\textbf{Temporal Sync.}}
& \multicolumn{2}{c}{\textbf{Audio Existence}}
& \multicolumn{2}{c}{\textbf{Sound Consistency}}
& \multirow{2}{*}{\textbf{Avg Gap}} \\
\cmidrule(lr){3-4} \cmidrule(lr){5-6} \cmidrule(lr){7-8}
&
& \textbf{Orig.} & \cellcolor{shiftbg}\textbf{\shift}
& \textbf{Orig.} & \cellcolor{mutebg}\textbf{\mute}
& \textbf{Orig.} & \cellcolor{swapbg}\textbf{\swap}
&  \\
\midrule
Gemini              & N/A  & 54.9        & \cellcolor{shiftbg}46.5 & 100.0 & \cellcolor{mutebg}13.4 & 93.6 & \cellcolor{swapbg}18.3 & \cellcolor{gapbg}56.8 \\
MiniCPM-o-4.5       & 9B   & 83.8        & \cellcolor{shiftbg}13.7 & 100.0 & \cellcolor{mutebg}19.0 & 95.8 & \cellcolor{swapbg}4.9  & \cellcolor{gapbg}\textcolor{red}{80.7} \\
Nemotron-3-Omni     & 30B  & 35.9        & \cellcolor{shiftbg}26.8 & 66.2  & \cellcolor{mutebg}4.2  & 88.7 & \cellcolor{swapbg}19.9 & \cellcolor{gapbg}46.6 \\
Qwen3-Omni          & 30B  & $100.0^{*}$ & \cellcolor{shiftbg}1.4  & 95.1  & \cellcolor{mutebg}0.0  & 75.4 & \cellcolor{swapbg}37.3 & \cellcolor{gapbg}77.3 \\
Ming-Omni-2.0       & 100B & 54.2        & \cellcolor{shiftbg}20.1 & 95.7  & \cellcolor{mutebg}54.9 & 90.1 & \cellcolor{swapbg}15.5 & \cellcolor{gapbg}49.8 \\
MiMo-V2.5           & 311B & 73.9        & \cellcolor{shiftbg}9.9  & 99.3  & \cellcolor{mutebg}2.1  & 89.4 & \cellcolor{swapbg}15.3 & \cellcolor{gapbg}78.4 \\
\bottomrule
\end{tabular}
\vspace{-0.3cm}
\end{table}

\subsection{Do Video-Capable Multimodal Models Rely on Visual Shortcuts?}
\label{sec:shortcut_diagnosis}

We examine whether video-capable multimodal models verify the audio stream or infer plausible sounds from visual context. \Cref{tab:shortcut_diagnosis} reports paired diagnostic accuracy under naturally correlated \textit{Original} controls and counterfactual interventions. Original videos serve as positive controls, while drops under \shift, \mute, or \swap reveal failures when natural audio-visual correlations are broken. \textbf{Avg Gap} measures the average accuracy drop from \textit{Original} to intervention conditions, with larger values indicating a larger performance collapse under counterfactual interventions. Its formula and the LLM-judge protocol for free-form outputs are provided in \Cref{app:prompts}.

\begin{wrapfigure}[20]{r}{0.58\textwidth}
\vspace{-12pt}
\centering
\includegraphics[width=\linewidth, trim=0 0 0 0, clip]{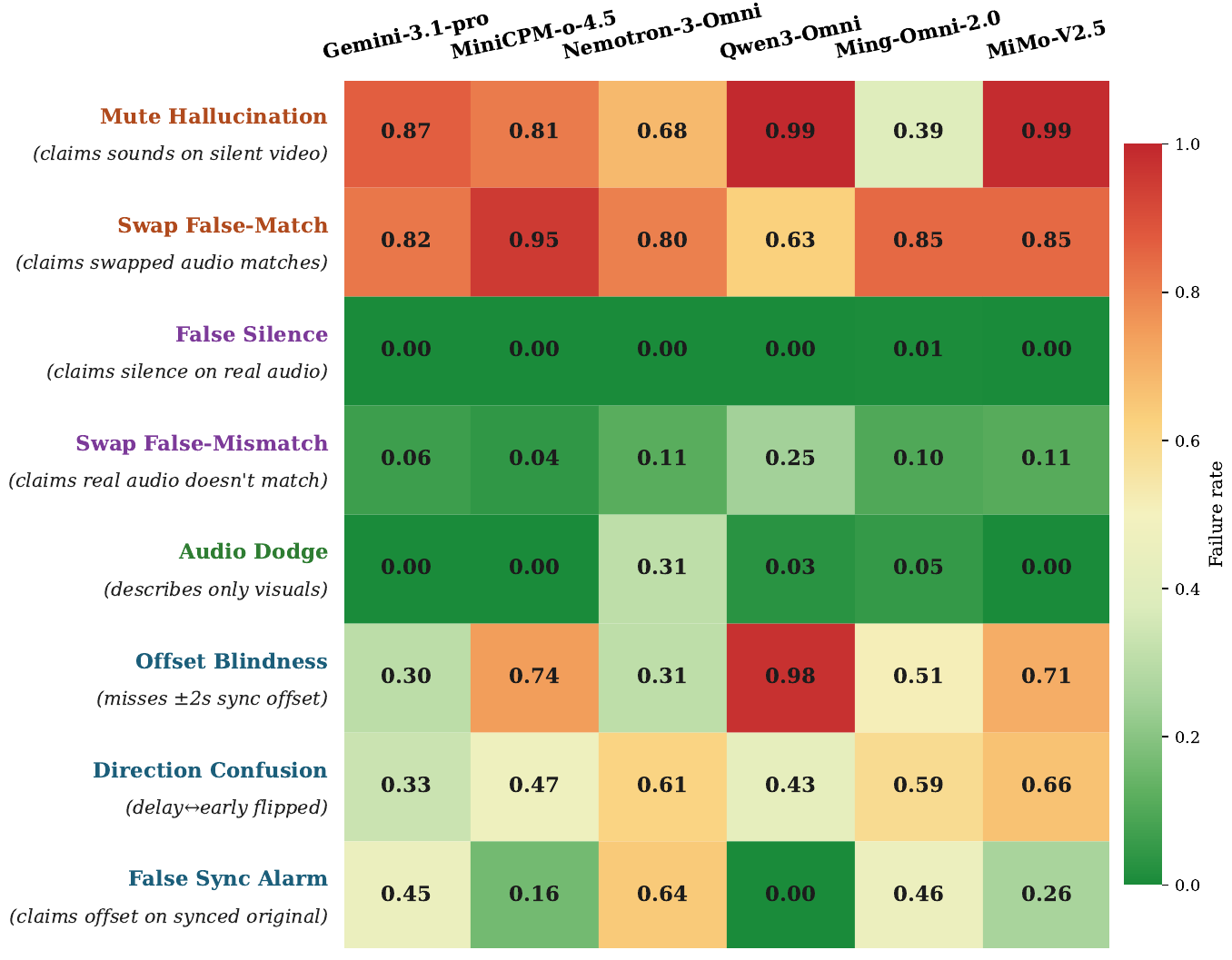}
\vspace{-1.5em}
\caption{
\textcolor{heatmap}{\textbf{Failure-mode heatmap}}.
Red indicates higher failure; audio hallucination dominates, while temporal failures are model-specific.
}
\label{fig:failure_heatmap}
\vspace{-0.5em}
\end{wrapfigure}

Overall, most models show large drops from \textit{Original} to intervention settings, indicating that strong performance on naturally correlated videos is fragile. MiniCPM-o-4.5 and MiMo-V2.5 have the largest gaps, 80.7\% and 78.4\%. Qwen3-Omni is diagnostic: its perfect original temporal-sync accuracy drops to 1.4\% under \shift, suggesting a synchronized-default prior rather than true temporal grounding. These results suggest that current models often rely on visual-semantic priors instead of verifying audio presence, timing, and source consistency.

\Cref{fig:failure_heatmap} exposes a uniform shortcut. Every model saturates on audio hallucination, with Mute Hallucination and Swap False-Match both above 0.63 across the board, while their symmetric counterparts (False Silence, Swap False-Mismatch) sit near zero: models invent audio that fits the visuals but rarely deny audio that is real. Temporal perception is worse. Qwen3-Omni misses 98\% of $\pm 2$\,s offsets; MiniCPM and MiMo miss roughly three quarters; and even when an offset \emph{is} flagged, the delay/early sign is wrong about half the time, close to a random label. Definitions for each axis are given in \Cref{app:failure-modes}.

\begin{figure}[b!]
\vspace{-1em}
    \centering
    \includegraphics[width=\linewidth]{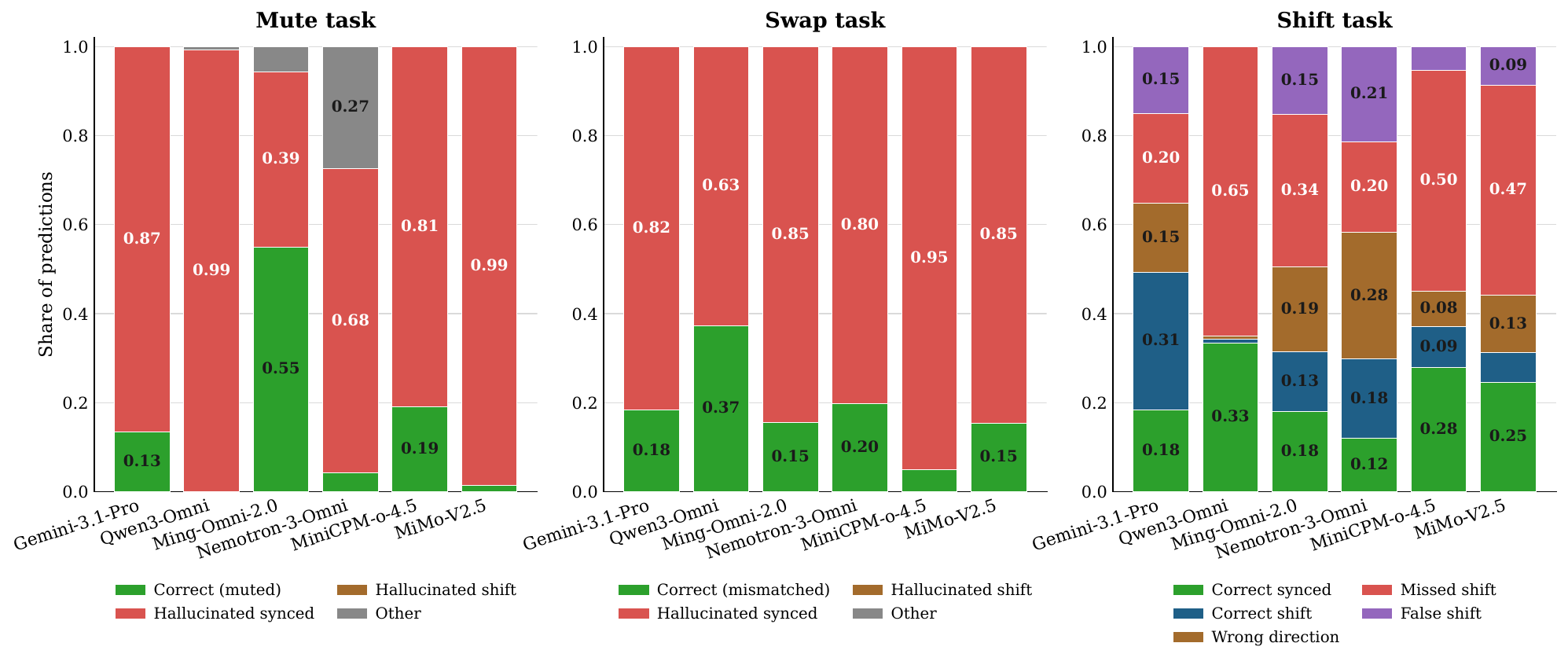}
    \vspace{-1.5em}
   \caption{
\textcolor{breakdownblue}{\textbf{Prediction breakdown}} per model on the three intervention tasks.
Errors cluster around a synced default, evidencing shortcut reliance over genuine audio-video alignment.
}
    \label{fig:breakdown}
    %\vspace{-1.6em}
    \vspace{-1em}
\end{figure}

\Cref{fig:breakdown} decomposes each model's predictions on the three intervention tasks. On Mute and Swap, almost all errors collapse onto Hallucinated synced, with five of six models fabricating matching audio on over 80\% of muted clips and the mismatched class recovered at most 37\% of the time. Hallucinated shift is negligible everywhere, indicating that models hold a strong \textit{synced} prior and rarely entertain temporal alternatives. The Shift panel makes the consequence concrete: Qwen3-Omni answers \textit{synced} on 98\% of inputs, while Gemini-3.1-Pro, Nemotron-3-Omni, and Ming-Omni-2.0 lose 19 to 22\% of predictions to Wrong direction, showing partial sensitivity to offsets without reliable sign recovery. Errors are systematically biased toward the synced prior rather than randomly distributed, indicating that current models rely on shortcut consistency rather than genuine cross-modal alignment.
\vspace{-1em}

\begin{table}[t!]
\vspace{-0.8em}
\caption{
Accuracy (\%) under different alignment recipes on temporal synchronization, general video and audio-visual understanding benchmarks. We evaluate temporal grounding on \textbf{\textcolor{syncfg}{Sync}} and \textbf{\textcolor{syncfg}{VGGSync}}, video understanding on \textbf{\textcolor{videofg}{V-MME}} and \textbf{\textcolor{videofg}{LVB}}, audio-visual understanding on \textbf{\textcolor{omnifg}{WS}} and \textbf{\textcolor{omnifg}{DO}}. \textbf{\textcolor{avgfg}{Avg.}} is the six-benchmark average. All DPO recipes are initialized from the SFT w/ OP checkpoint.
}
\vspace{3pt}

\label{tab:alignment_tax}
\centering
\small
\setlength{\tabcolsep}{5.5pt}
\renewcommand{\arraystretch}{1.05}
\begin{tabular}{lSSVVOOA}
\toprule
\textbf{Recipe}
& \multicolumn{1}{c}{\textbf{\textcolor{syncfg}{Sync}}}
& \multicolumn{1}{c}{\textbf{\textcolor{syncfg}{VGGSync}}}
& \multicolumn{1}{c}{\textbf{\textcolor{videofg}{V-MME}}}
& \multicolumn{1}{c}{\textbf{\textcolor{videofg}{LVB}}}
& \multicolumn{1}{c}{\textbf{\textcolor{omnifg}{WS}}}
& \multicolumn{1}{c}{\textbf{\textcolor{omnifg}{DO}}}
& \multicolumn{1}{c}{\textbf{\textcolor{avgfg}{Avg.}}} \\
\midrule
Qwen3-Omni-30B
& 34.3 & 36.8 & 69.2 & 49.1 & 50.3 & 68.2 & 51.3 \\
% MiniCPMo-4.5
% & 37.1 & 33.4 & 67.9 & 47.3 & \textbf{52.3} & \textbf{71.0} & 51.5 \\
SFT w/ OP
& 73.9 & -- & -- & -- & -- & -- & -- \\
SFT w/ CTP + FV-D + FV-AL
& 76.1 & 46.7 & 43.8 & 40.8 & 48.2 & 66.9 & 53.8 \\
\midrule
DPO w/ SP
& 75.4 & 55.7 & 69.3 & 50.9 & 49.8 & \textbf{69.0} & 61.7 \\
DPO w/ OP + SP
& 76.5 & 56.4 & 69.9 & 47.7 & 49.7 & 68.5 & 61.5 \\
DPO w/ SP + FV-D
& 82.2 & 55.4 & 69.1 & 51.5 & 49.8 & 68.0 & 62.7 \\
DPO w/ OP + FV-D + LV-MCQA
& 83.0 & \textbf{56.6} & 69.2 & 50.4 & 49.9 & 67.6 & 62.8 \\
DPO w/ CTP + FV-D
& 81.2 & 55.8 & 69.6 & 51.4 & 49.5 & 68.0 & 62.6 \\
DPO w/ CTP + FV-D + LV-MCQA
& 82.2 & 55.7 & 69.2 & 51.1 & 49.8 & 67.8 & 62.6 \\
DPO w/ CTP + FV-D + FV-A
& 82.6 & 55.9 & 69.1 & 50.8 & 49.9 & 67.3 & 62.6 \\
\midrule
\textbf{Ours}
& \textbf{83.1} & 56.4 & \textbf{70.1} & \textbf{52.1} & \textbf{50.3} & 67.9 & \textbf{63.3} \\
\bottomrule
\end{tabular}

\vspace{0.2em}
\footnotesize{
OP: initial original-sync preference data; SP: SFT-policy negatives; CTP: counterfactual temporal preferences; FV-* and LV-MCQA denote general video preference data.
}
\vspace{-0.5cm}
\end{table}

\subsection{Targeted Alignment Improves Temporal Grounding Without Alignment Tax}
\label{sec:alignment_tax}

\begin{wrapfigure}[18]{r}{0.60\textwidth}
\vspace{-12pt}
\centering
\includegraphics[width=\linewidth, trim=0 0 0 0, clip]{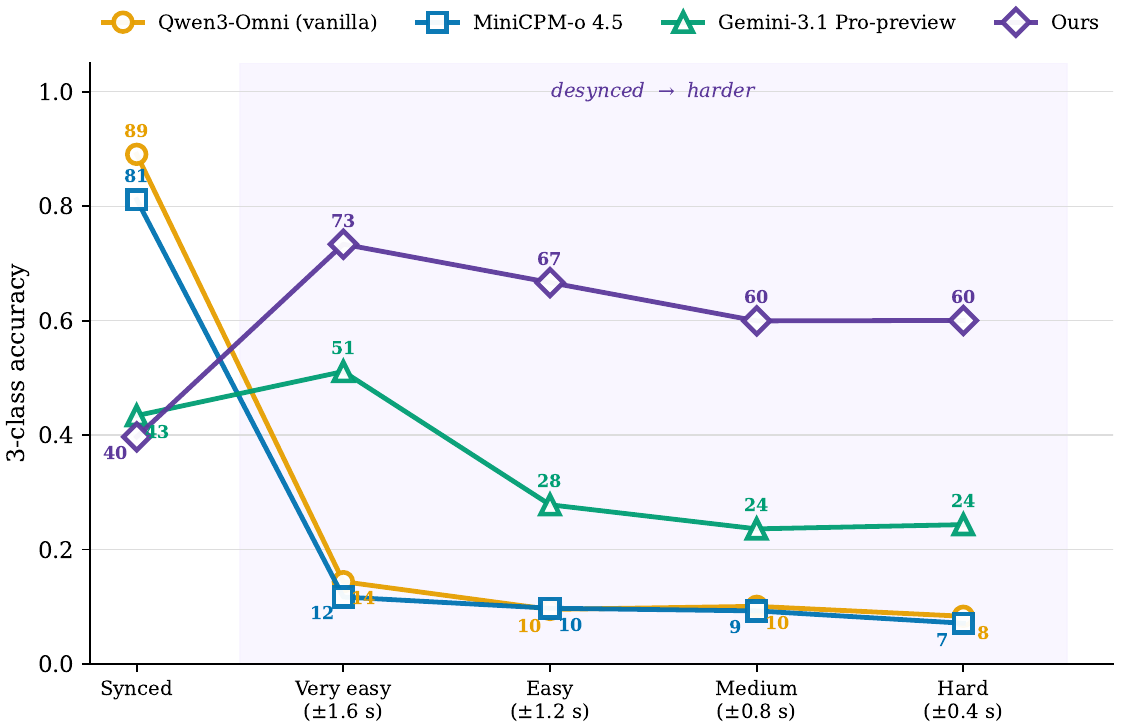}
\vspace{-1.5em}
\caption{
\textcolor{difficulty}{\textbf{Difficulty-band robustness}}.
Smaller offsets are harder; our model remains robust while baselines collapse under desynchronization.
}
\label{fig:vgg_diff}
\vspace{-4pt}
\end{wrapfigure}

We next ask whether targeted intervention training can improve temporal grounding without hurting general capabilities. Starting from Qwen3-Omni-30B, we compare alignment recipes using original synchronization preferences, self-sampled negatives, counterfactual temporal preferences, and general video preferences. \textbf{Ours} denotes our final 10K DPO recipe combining CTP, FV-D, and FV-A-L.~\Cref{app:recipe-data} details each data source, including its construction, preference format, and intended training signal.

\Cref{tab:alignment_tax} shows that alignment training substantially improves temporal synchronization over the vanilla Qwen3-Omni baseline. Our best 10K mixture improves \textbf{\textcolor{syncfg}{Sync}} from 34.3\% to 83.1\% and \textbf{\textcolor{syncfg}{VGGSync}} from 36.8\% to 56.4\%, suggesting that the model gains transferable temporal grounding rather than simply memorizing our intervention format. At the same time, it maintains or improves \textbf{\textcolor{videofg}{V-MME}}, \textbf{\textcolor{videofg}{LVB}}, and \textbf{\textcolor{omnifg}{WS}}, remains competitive on \textbf{\textcolor{omnifg}{DO}}, and raises the six-benchmark average accuracy from 51.3\% to 63.3\%. The contrast with the SFT-only mixture, which improves \textbf{\textcolor{syncfg}{Sync}} but sharply hurts general benchmarks, indicates that preference alignment rather than supervised mixing is key to improving temporal grounding without incurring an alignment tax.

The recipe ablation further clarifies which data sources are responsible for this tradeoff. SFT with intervention and general video data already improves \textbf{\textcolor{syncfg}{Sync}}, but substantially degrades \textbf{\textcolor{videofg}{V-MME}} and \textbf{\textcolor{videofg}{LVB}}, indicating that supervised mixing alone can over-specialize the model to intervention-style supervision. In contrast, DPO recipes recover general capability while preserving temporal gains. Self-sampled preferences provide a strong general baseline, but the best temporal results arise when targeted temporal preferences are combined with general video preference data. This suggests that counterfactual temporal supervision supplies the grounding signal, while FineVideo and LLaVA-Video preferences regularize the model toward broad video understanding.

\begin{figure*}[t!]
    \centering
    \vspace{-1em}
    \begin{subfigure}[t]{0.495\textwidth}
        \centering
        \includegraphics[width=\linewidth]{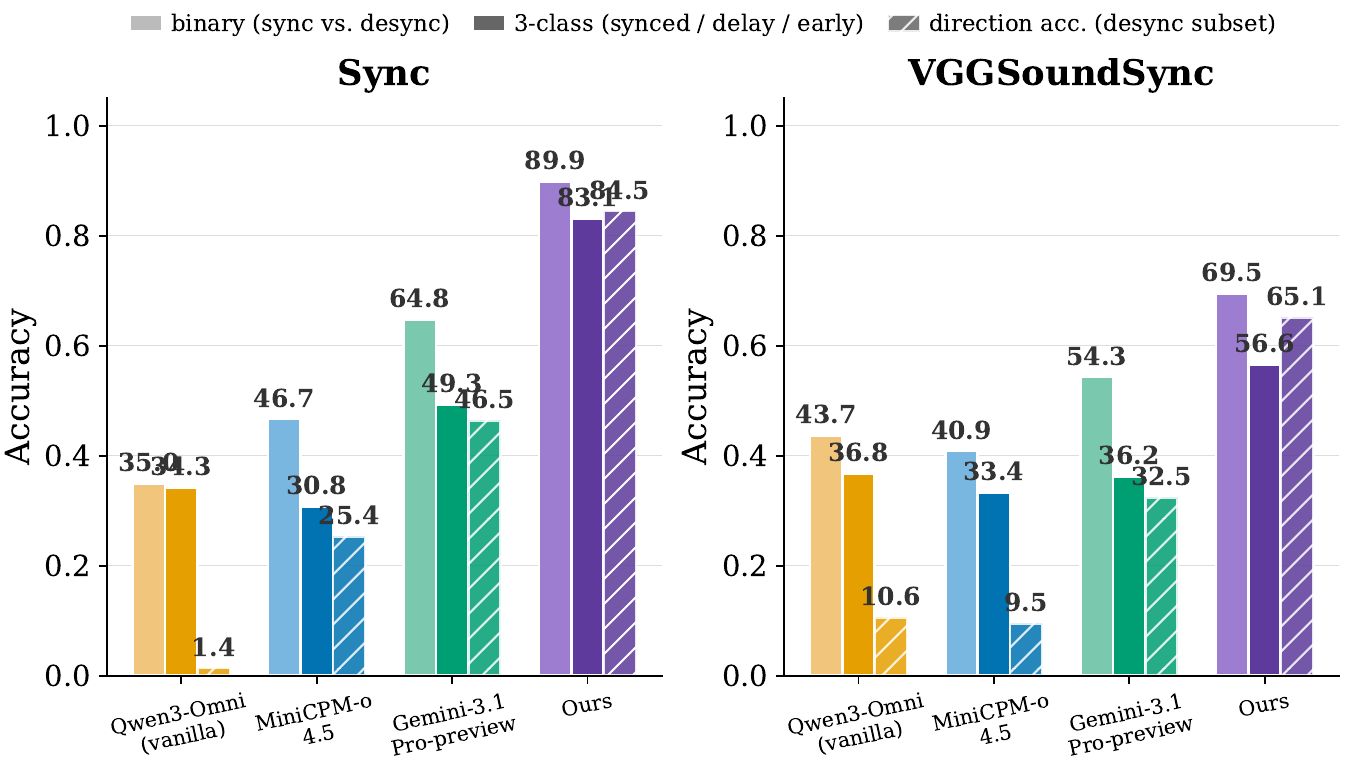}
        \caption{Audio-visual synchronization accuracy.}
        \label{fig:headline-accuracy}
    \end{subfigure}
    \hfill
    \begin{subfigure}[t]{0.495\textwidth}
        \centering
        \includegraphics[width=\linewidth]{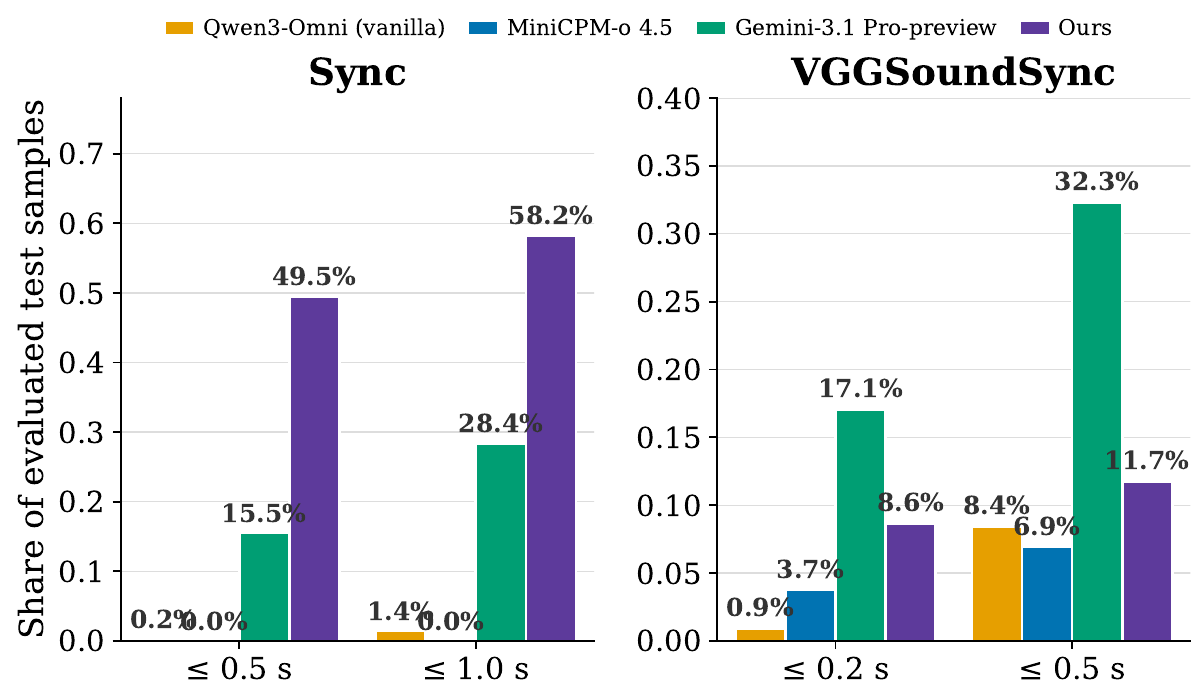}
        \caption{Localization quality under offset tolerance.}
        \label{fig:offset-coverage}
    \end{subfigure}
    %\vspace{-0.5em}
    \caption{\textcolor{verifyfg}{\textbf{Complementary synchronization results.}} Left: model accuracy on binary synchronization, three-way temporal classification, and direction prediction. Right: the fraction of samples whose predicted offset is close to the ground-truth temporal displacement.}
    \label{fig:sync-results-combined}
    %\vspace{-0.8cm}
    \vspace{-1em}
\end{figure*}

\Cref{fig:vgg_diff} evaluates synchronization across temporal-offset difficulty bands on \textbf{\textcolor{syncfg}{VGGSync}}, using the \shift intervention from \Cref{sec:dspi}. Each band corresponds to a different offset magnitude $|\Delta|$. The high synced accuracy of vanilla Qwen3-Omni and MiniCPM-o should be read together with \Cref{fig:breakdown}: both models strongly prefer answering ``synced,'' making them appear accurate only when no shift is applied. Once any nonzero offset is introduced, their accuracy collapses across all bands, including large $|\Delta|$ values that should be easy to detect. Gemini-3.1-Pro follows a more expected trend, performing better on larger shifts and degrading as $|\Delta|$ becomes smaller and subtler. Our model remains stronger across all shifted bands while also reflecting the expected pattern that smaller $|\Delta|$ is harder. This suggests that temporal grounding should be judged not by synced-video accuracy alone, but by whether models show difficulty-sensitive verification under controlled audio displacement.

\begin{wrapfigure}[14]{r}{0.63\textwidth}
\vspace{-14pt}
\centering
\includegraphics[width=\linewidth, trim=0 0 0 0, clip]{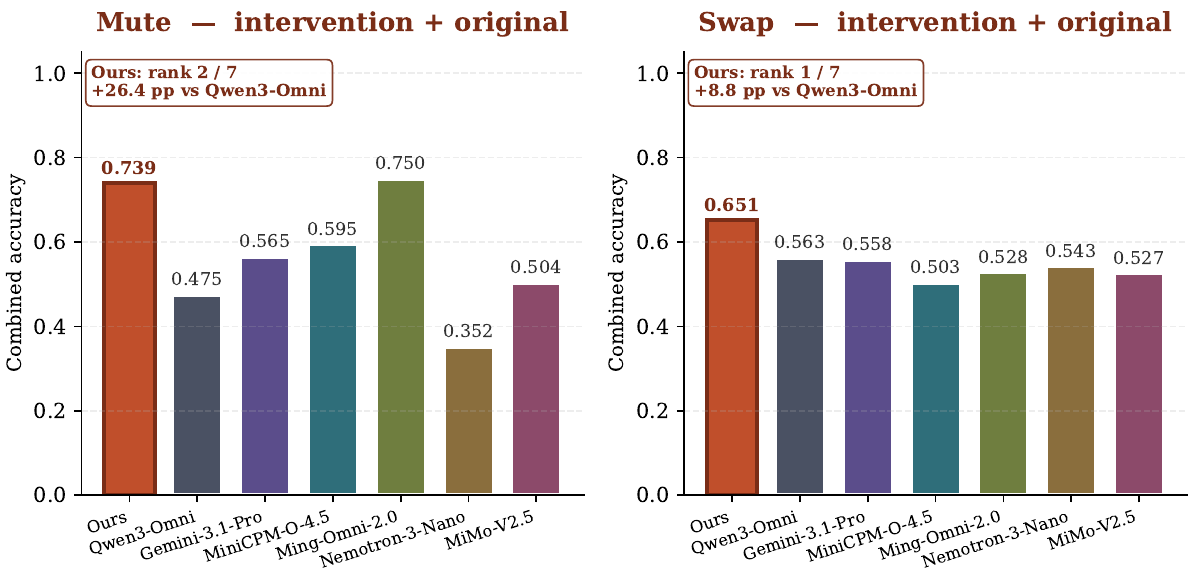}
\vspace{-1.5em}
\caption{
\textcolor{oursfg}{\textbf{Beyond temporal synchronization}}.
Combined Mute and Swap accuracy over original and intervened conditions.
}
\label{fig:beyond_sync}
\vspace{-4pt}
\end{wrapfigure}

\Cref{fig:sync-results-combined} separates temporal grounding into label-level synchronization detection and fine-grained offset localization. In \Cref{fig:headline-accuracy}, our model consistently outperforms Gemini-3.1-Pro across all synchronization metrics, including binary synced/desynced classification, three-way temporal classification, and direction prediction on desynced videos. This suggests that the improvement is not limited to coarse mismatch detection, but extends to the harder problem of identifying the temporal direction of the mismatch. \Cref{fig:offset-coverage} further sharpens this distinction: most baselines rarely predict offsets close to the ground truth, whereas our model achieves the strongest localization coverage on \textbf{\textcolor{syncfg}{Sync}} and remains competitive on \textbf{\textcolor{syncfg}{VGGSync}}. Together, these results show that audio-visual grounding should not be measured only by whether a model flags desynchronization, but also by whether it can localize the temporal mismatch with meaningful precision.

\vspace{-3pt}
\subsection{Beyond Temporal Synchronization}\label{sec:beyond_temporal}
\Cref{fig:beyond_sync} evaluates whether the recipe in \Cref{tab:alignment_tax} can extend beyond temporal synchronization. Starting from our best recipe, we add a small amount of Mute/Swap SFT. The resulting model ranks first on Swap and second on Mute, yielding a 28\% average gain over vanilla Qwen3-Omni across Shift, Mute, and Swap. \Cref{fig:falsealarm} further separates intervention detection from false alarms on original controls, showing that the gain is not merely higher combined accuracy: our model moves closer to the ideal top-left tradeoff, especially on Swap. This suggests that intervention-based training can mitigate multiple shortcut modes, while audio existence and cross-modal consistency still require targeted supervision beyond temporal alignment alone.

%% file: sections/02_Related_Work.tex
\section{Related Work}

\begin{wrapfigure}[14]{r}{0.60\textwidth}
\vspace{-10pt}
\centering
\includegraphics[width=\linewidth, trim=0 0 0 0, clip]{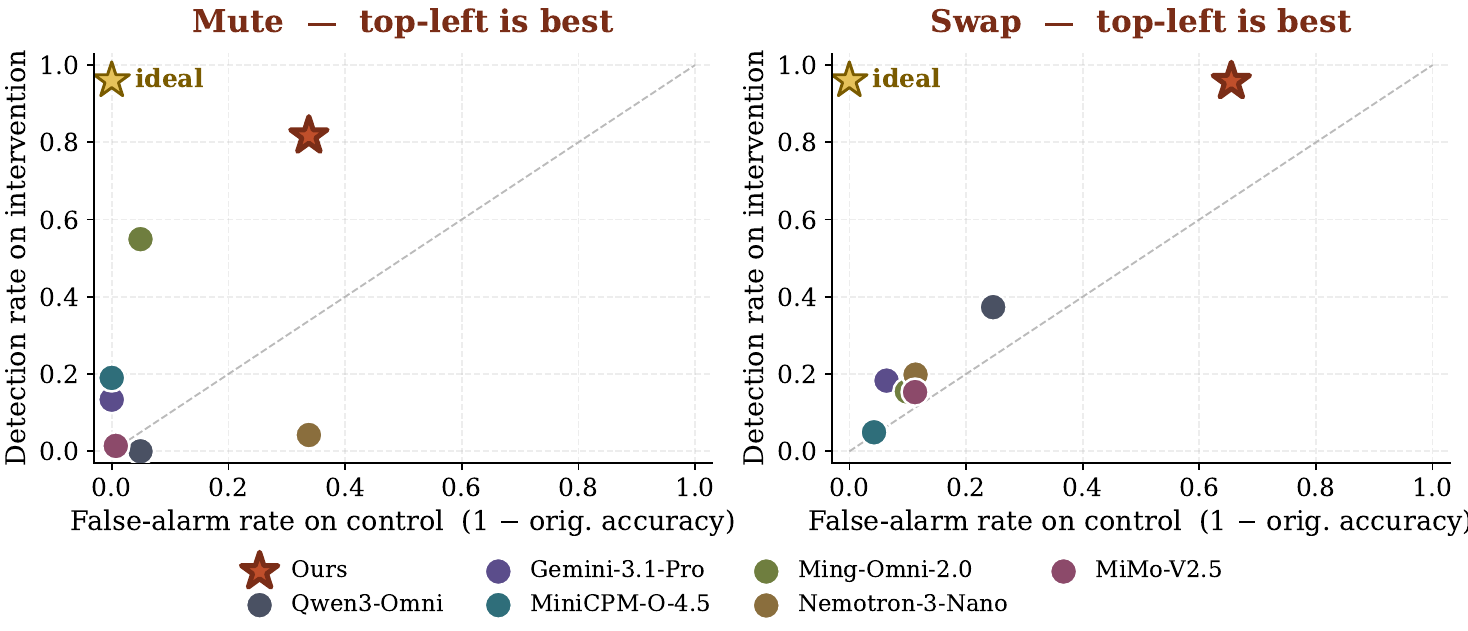}
\caption{
\textcolor{tradeofffg}{\textbf{Intervention-control tradeoff}}.
Top-left indicates strong intervention detection with few false alarms on original controls.
}
\label{fig:falsealarm}
\vspace{-0.5cm}
\end{wrapfigure}

\paragraph{Native Omni Models and Cross-Modal Shortcuts}
Recent frontier multimodal models are shifting from frame-centric video-language pipelines toward native multimodal or omni-modal processing, where video, audio, images, and text are handled through a unified interface or architecture~\cite{hurst2024gpt,qwenteam2026qwen35omnitechnicalreport,li2025baichuan}. Although such integration suggests stronger audio-visual grounding~\cite{wu2024nextgpt,zhan2024anygpt,Girdhar2023ImageBindOE}, it does not ensure that models verify the audio stream. The shortcut behavior we observe reflects a long-standing assumption in audio-visual representation learning: natural videos provide supervision because visual and acoustic events often co-occur in synchronized and semantically aligned ways~\cite{Korbar2018CooperativeLO,Morgado2020AudioVisualID,Chen2020VggsoundAL,Senocak2018LearningTL,Alwassel2019SelfSupervisedLB}. While effective for learning shared representations, these co-occurrence signals can conflate genuine grounding with statistical association~\cite{Morgado2021RobustAI,Yuksekgonul2022WhenAW,Thrush2022WinogroundPV}. Models may therefore rely on visual-semantic shortcuts~\cite{agrawal-etal-2016-analyzing,rohrbach-etal-2018-object}: barking dogs imply barks, falling objects imply impacts, and speaking faces imply speech. Without negative cases that break these correlations~\cite{Singh2023LookingSS}, models can appear grounded without checking whether sound is present, synchronized, or physically consistent, producing a \textit{Clever Hans} effect~\cite{Lapuschkin2019UnmaskingCH} in modern audio-visual models~\cite{Buch2022RevisitingT,Wang2024VideoHallucerEI}. We address this gap using controlled audio interventions that test cross-modal verification under broken audio-visual correlations.

\paragraph{Preference Alignment for Video-Capable Multimodal Models}
Video-capable multimodal models have evolved along two related directions: video-language instruction tuning, which connects visual encoders with LLMs for video understanding~\cite{DBLP:conf/emnlp/LinYZCNJ024,DBLP:conf/nips/Dai0LTZW0FH23}, and native omni-modal modeling, which integrates video, audio, images, and text within unified interfaces or architectures~\cite{hurst2024gpt,qwenteam2026qwen35omnitechnicalreport,li2025baichuan}. Preference-based methods such as Direct Preference Optimization~\cite{rafailov2023direct} have also been adapted to video-language modeling, often using detailed captions or language-model feedback as proxies for video-grounded rewards~\cite{zhang-etal-2025-direct}. However, existing alignment data mainly targets helpfulness~\cite{DBLP:conf/nips/Ouyang0JAWMZASR22,Bai2022TrainingAH}, visual question answering~\cite{NEURIPS2023_6dcf277e}, instruction following~\cite{wei2022finetuned}, and safety~\cite{Zong2024SafetyFA,Zhu2025OmniGuardUO}, with limited attention to how models use, ignore, or misattribute the audio stream. Recent work has also observed visual dominance and video-driven audio hallucination in audio-visual LLMs~\cite{selvakumar2026audio,baid2026don}. Our work instead decomposes audio-visual grounding into temporal synchronization, audio existence, and cross-modal material consistency, and studies how intervention data and preference optimization affect each dimension.

%% file: sections/05_Conclusion.tex
\vspace{-0.5em}
\section{Conclusion}

This work shows that apparent audio understanding in video-capable multimodal models can be strongly vision-driven. We identify this behavior as an audio-visual \textit{Clever Hans} effect, where models answer sound-related questions by exploiting natural visual-acoustic correlations rather than verifying the observed audio stream. To make this failure measurable, we introduce \Thud, which uses \shift, \mute, and \swap interventions to probe temporal synchronization, sound existence, and audio-visual consistency. Our experiments reveal systematic shortcut reliance across current open and closed models. We further show that counterfactual intervention data can be used not only for diagnosis, but also for alignment: a two-stage recipe combining intervention-derived preferences with event-level general video preferences improves audio-visual grounding while preserving broad video understanding. Overall, our findings suggest that future video-capable models should be evaluated and trained under counterfactual audio-visual conditions, not only naturally correlated videos.

%% file: sections/Appendix.tex
\crefalias{section}{appendix}
\crefalias{subsection}{appendix}
\crefalias{subsubsection}{appendix}

\section{Schematic Overviews of Data Construction and Alignment }
\subsection{Data Construction Pipeline}
\label{app:data-construction-pipeline}
\begin{figure}[t]
    \centering
    \includegraphics[width=\linewidth]{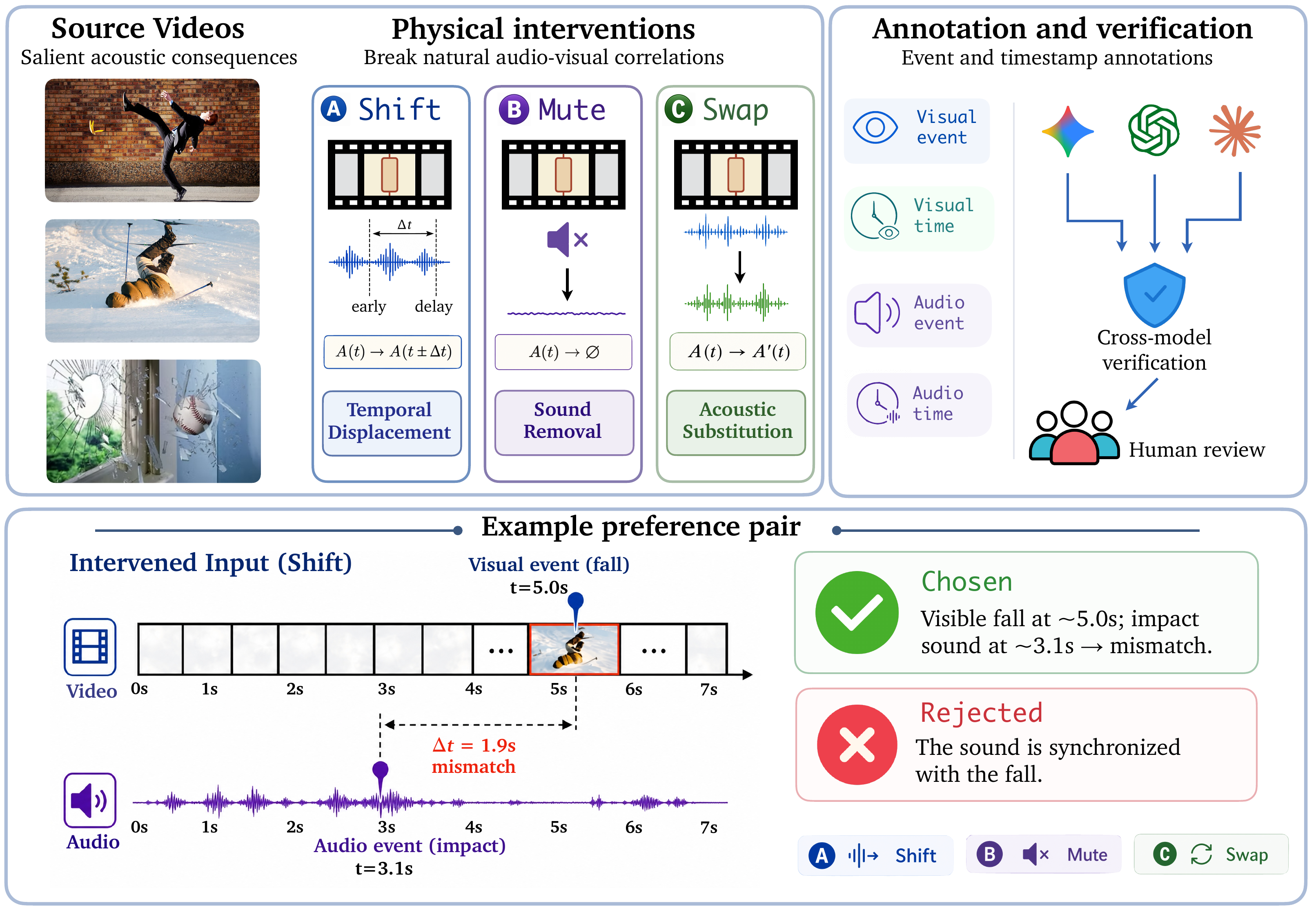}
    \caption{
Pipeline for intervention \textcolor{pipelineblue}{\textbf{data construction}}. We create \shift, \mute, and \swap variants from source videos with salient acoustic events, annotate visual/audio events and timestamps via cross-model verification with human review, and construct chosen--rejected preference pairs for training. The bottom panel shows a representative \shift example.
}
    \label{fig:data-construction}
    \vspace{-0.3cm}
\end{figure}

\Cref{fig:data-construction} illustrates the systematic pipeline for constructing the intervention-driven preference dataset. The process begins with initial event-time labeling using Gemini, which are then rigorously cross-verified: visual timestamps are validated through a consensus of GPT and Claude via frame-unit analysis, while acoustic timestamps undergo human inspection to ensure ground-truth reliability. 
After filtering samples based on strict agreement criteria, we apply three interventions, \shift, \mute, and \swap, to the validated source videos. This results in the final preference pairs, where the "chosen" response reflects true audio-visual grounding and the "rejected" response exposes the visually-plausible shortcuts we aim to mitigate during alignment.

\subsection{Intervention Summary}
\label{app:intervention_summary}
\Cref{tab:intervention-summary} summarizes the three interventions in \Thud. 
Each intervention keeps the visual stream fixed while perturbing the audio track to target one grounding dimension: \shift probes temporal synchronization, \mute probes sound existence, and \swap probes source consistency. 
These controlled cases test whether models verify the observed audio or simply infer plausible sounds from visual priors.

\subsection{Preference Data Sources}
\label{app:recipe-data}

This section describes the preference data sources used in our alignment recipe study. Each preference example is represented as a pair of responses, where the chosen response provides the desired behavior and the rejected response provides a shortcut-prone or incorrect alternative.

\paragraph{Original synchronization preferences (OP).}
Original synchronization preferences are constructed from the annotated audio-visual event tuples introduced in \Cref{sec:AAPC}. For each video, we represent the aligned visual and acoustic event as
\begin{equation}
    z_i = (e_i^v, t_i^v, e_i^a, t_i^a),
\end{equation}
where $e_i^v$ and $t_i^v$ denote the visual event and its timestamp, while $e_i^a$ and $t_i^a$ denote the corresponding acoustic event and timestamp. The chosen response is the annotated answer derived from the original aligned event. The rejected response is produced by perturbing one or more components of $z_i$, such as the visual event, visual timestamp, acoustic event, or acoustic timestamp, creating a plausible but incorrect synchronization explanation.

\paragraph{SFT-policy negatives (SP).}
Self-sampled negatives are generated from the SFT model itself. Given the same video-question input, we use the reference annotation as the chosen response and treat the SFT model's incorrect or shortcut-prone output as the rejected response. This data source encourages the model to correct its own post-SFT failure modes.

\paragraph{Counterfactual temporal preferences (CTP).}
Counterfactual temporal preferences are constructed by pairing original and shifted videos. For an original video, the chosen response corresponds to the original synchronized condition, while the response describing the shifted condition is used as the rejected response. For a shifted video, this assignment is reversed: the shifted-condition answer is chosen, and the original synchronized answer is rejected. This forces the model to distinguish true temporal alignment from visually plausible but temporally inconsistent audio.

\paragraph{FineVideo descriptive preferences (FV-D).}
FV-D is derived from the FineVideo data described in \Cref{app:finevideo-data}. We use the description, localization, and attribution tasks, which encourage the model to produce faithful video descriptions, localize relevant events, and attribute answers to appropriate visual or audio evidence.

\paragraph{FineVideo audio-visual QA preferences (FV-AVQA).}
FV-AVQA corresponds to the audio-dependent QA subset in \Cref{app:finevideo-data}. These examples ask questions that require audio evidence. Candidate questions and answers are generated by Gemini and then manually filtered. We further retain examples where GPT-based text-only answering fails, since GPT does not receive the audio stream. This filtering emphasizes cases where the answer cannot be reliably inferred from visual or textual priors alone.

\paragraph{FineVideo audio-visual QA long-form preferences (FV-AVQA-L).}
FV-AVQA-L is a long-form version of FV-AVQA. Instead of only selecting an answer option, the chosen response includes both the answer and an explanation grounded in the audio-visual evidence. This data source encourages the model to justify audio-dependent answers rather than relying on short-form guesses.

\paragraph{LLaVA-Video multiple-choice QA (LV-MCQA).}
LV-MCQA is a multiple-choice video QA dataset titled LLaVA-Video-178K ~\cite{zhang2024videoinstructiontuningsynthetic}. We include it as a general video preference source to regularize the model toward broad video understanding and reduce over-specialization to intervention-style examples.

\subsection{Alignment Pipeline}
\label{app:alignment-pipeline}
\begin{figure}[t]
    \centering
    \includegraphics[width=\linewidth]{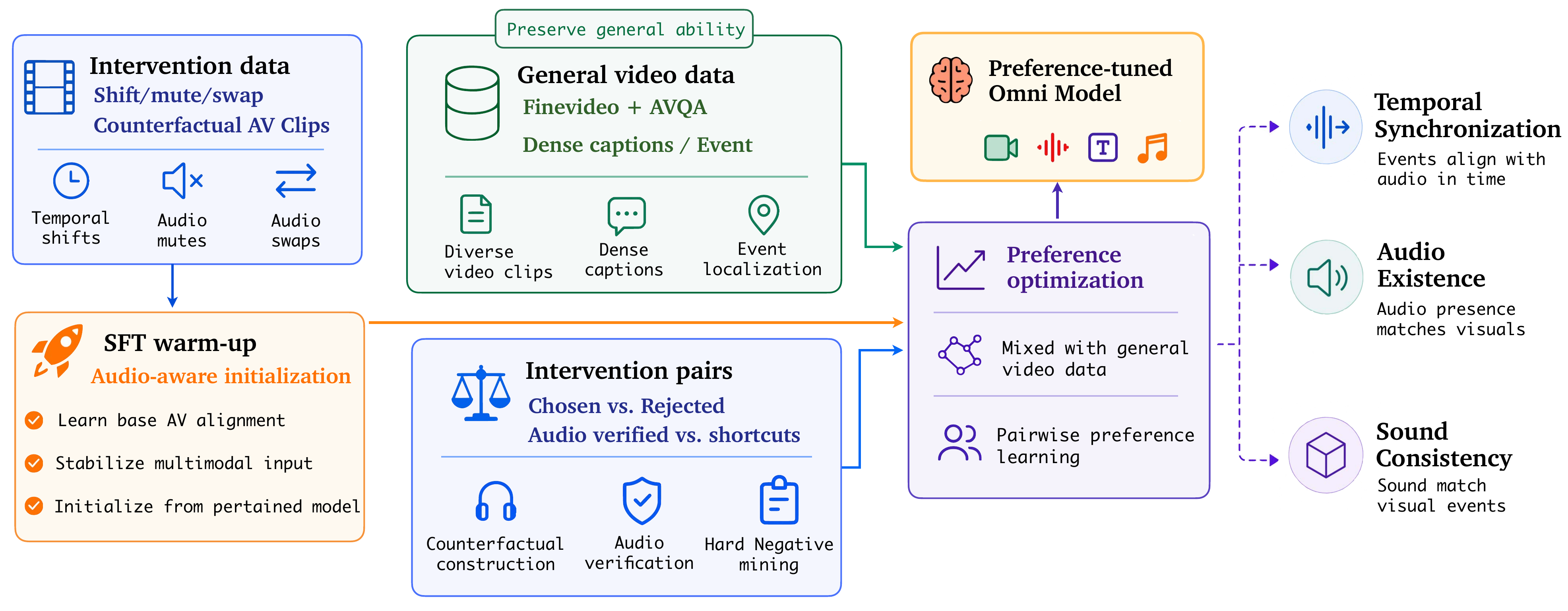}
   \caption{
Two-stage intervention-driven \textcolor{alignorange}{\textbf{alignment pipeline}}.
Counterfactual intervention data is first used for SFT warm-up, and intervention preference pairs are then mixed with general video data during preference optimization.
This design encourages audio-verified responses while preserving general video understanding.
}
    \label{fig:preference_optimization}
    \vspace{-1.0em}
\end{figure}

\Cref{fig:preference_optimization} illustrates our two-stage post-training pipeline designed to detect \shift, \mute, and \swap failures while preserving general video understanding. 
The training process integrates our targeted intervention dataset and re-annotated general video instructions derived from FineVideo~\cite{Farre2024FineVideo}. 
In Stage 1, an SFT warm-up on intervention data establishes basic audio-aware patterns. Stage 2 applies DPO using a mixture of intervention preference pairs and general video data. The preference pairs teach the model to reject visually plausible shortcuts, while the general data acts as a regularizer to preserve broad multimodal capabilities.

\begin{table}[h]
\caption{
Summary of our three physical interventions. Each intervention breaks a different natural audio-visual correlation and poses a diagnostic question about audio-visual grounding.
}
\label{tab:intervention-summary}
\centering
\begin{tabular}{lccc}
\toprule
\textbf{Intervention} 
& \textbf{Operation} 
& \textbf{Broken correlation}
& \textbf{Diagnostic question} \\
\midrule
\shift
& $a_{1:T} \rightarrow a_{1:T}^{+\Delta}$ 
& temporal synchrony
& Is the sound synchronized? \\
\mute 
& $a_{1:T} \rightarrow \varnothing$ 
& sound existence
& Is any sound present? \\
\swap
& $a_{1:T} \rightarrow a'_{1:T}$ 
& source consistency
& Does the sound match the event? \\
\bottomrule
\end{tabular}
\vspace{-0.5em}
\end{table}

%============================================
\section{Annotation and Verification Details}
\label{app:annotation}

\paragraph{Video-to-frame-unit conversion.}
For visual verification with GPT and Claude, we convert each video into $N$ temporally ordered frame units. Given a video of duration $T$ seconds, we split it into non-overlapping windows
\[
    u_j = [s_j, e_j], \qquad j=1,\ldots,N,
\]
where $s_j$ and $e_j$ denote the start and end time of the $j$-th unit. From each unit, we sample representative frames and present them in temporal order, together with the timestamp range of the unit. The verifier is asked to select the unit that contains the target visual event and optionally refine the timestamp within that unit. This frame-unit format allows models without direct video ingestion to perform temporal localization over visual evidence.

\paragraph{Gemini annotation prompt.}
We use Gemini to produce the initial audio-visual event annotation. The prompt is designed to avoid generic captioning and instead force event-level localization:
\begin{tcolorbox}[colback=gray!3,colframe=gray!35,boxrule=0.5pt,arc=2pt]
\small
You are given a video with audio. Identify the most salient visible event that has a corresponding acoustic consequence.  
Return a JSON object with the following fields:

\texttt{visual\_event}: a short description of the visible event.  

\texttt{visual\_time}: the timestamp in seconds when the visible event occurs.  

\texttt{audio\_event}: a short description of the corresponding sound.  

\texttt{audio\_time}: the timestamp in seconds when the sound occurs.  

\texttt{confidence}: high / medium / low.

If the visual event or audio event cannot be localized reliably, return \texttt{uncertain}.
\end{tcolorbox}

\paragraph{Frame-unit visual verification prompt.}
For GPT and Claude, we provide the ordered frame units and the candidate visual event proposed by Gemini:
\begin{tcolorbox}[colback=gray!3,colframe=gray!35,boxrule=0.5pt,arc=2pt]
\small
You are given temporally ordered frame units from a video. Each unit contains representative frames and a timestamp range.  
Target visual event: \texttt{[visual\_event]}.

Select the frame unit where this event occurs. 

Return:
\texttt{unit\_id}, \texttt{timestamp\_range}, and a one-sentence justification.  
If the event is not visible or cannot be localized, return \texttt{uncertain}.
\end{tcolorbox}

\paragraph{Agreement and filtering rules.}
We retain a sample only if it satisfies the following conditions:
\begin{enumerate}[leftmargin=1.5em,itemsep=2pt,topsep=2pt]
    \item \textbf{Visual agreement:} Gemini, GPT, and Claude localize the visual event within $\epsilon_v = 0.8$ seconds, or select overlapping frame units.
    \item \textbf{Audio verification:} the acoustic event is audible and its timestamp can be verified by human inspection within $\epsilon_a = 0.5$ seconds of the Gemini prediction.
    \item \textbf{Event clarity:} the visual event has a clear onset or peak moment, such as impact, fall, collision, breakage, or contact.
    \item \textbf{Acoustic salience:} the corresponding sound is not dominated by unrelated background music, speech, or noise.
    \item \textbf{Intervention validity:} after applying \shift, \mute, or \swap, the correct answer remains unambiguous.
\end{enumerate}

\paragraph{Manual review protocol.}
Samples failing automatic agreement are manually reviewed. We use the following decision rules. If the disagreement is due to a small boundary ambiguity, we correct the timestamp to the clearest event onset. If the visual event is partially occluded, spread over a long interval, or lacks a well-defined moment, we discard the sample. If the sound is too weak, masked by background noise, or not clearly tied to the visual event, we discard the sample. For \swap, we additionally discard cases where the substituted audio is trivially unrelated or too similar to the original audio; retained swaps must be acoustically plausible but inconsistent with the visible event.

% =============================================

\section{Experimental Configuration}
\label{app:experimental-configuration}

This section provides additional implementation details for our supervised fine-tuning (SFT), preference optimization, and evaluation experiments. All training experiments are conducted on 8*NVIDIA H200 GPUs, while evaluation experiments are conducted on either 8*NVIDIA H200 or 8*NVIDIA H100 GPUs. A single SFT run takes approximately 6 hours, and DPO training on 10K examples takes approximately 20 hours. For evaluation, the average inference time across the six datasets is approximately 5 hours per dataset.

\paragraph{Base model.}
We use Qwen3-Omni-30B-A3B-Instruct as the base omni-modal model. Video inputs are processed with audio enabled by setting \texttt{use\_audio\_in\_video=true}.

\paragraph{Training configuration.}
We summarize the training configurations for supervised fine-tuning and preference optimization in \Cref{tab:training-config}. Both stages are launched with \texttt{torchrun} on a single node with 8 GPUs, using DeepSpeed ZeRO-3 for memory-efficient distributed training. 

\begin{table}[t]
\centering
\caption{Training configurations for the supervised fine-tuning and DPO stages.}
\vspace{3pt}
\label{tab:training-config}
\footnotesize
\setlength{\tabcolsep}{5pt}
\renewcommand{\arraystretch}{1.12}
\begin{tabular}{p{0.24\linewidth}p{0.34\linewidth}p{0.34\linewidth}}
\toprule
\textbf{Configuration} & \textbf{SFT} & \textbf{DPO} \\
\midrule
Initialization 
& Qwen3-Omni-30B-A3B-Instruct 
& SFT checkpoint \\
Fine-tuning type 
& Full-parameter tuning 
& LoRA \\
Epochs 
& 3 
& 1 \\
\midrule
Learning rate 
& $2\times 10^{-6}$ 
& $1\times 10^{-6}$ \\
Scheduler / warmup 
& Cosine / 0.03 
& Cosine / 0.03 \\
Weight decay / grad norm 
& 0.01 / 1.0 
& 0.0 / 1.0 \\
Precision 
& bf16 
& bf16 \\
\midrule
Cutoff length 
& 131,072 
& 131,072 \\
Video max pixels 
& 501,760 
& 250,880 \\
Audio in video 
& Enabled 
& Enabled \\
\midrule
Batch size 
& 1 per GPU; accum. 4; effective 32 
& 1 per GPU; accum. 8; effective 64 \\
Memory optimization 
& DeepSpeed ZeRO-3 
& DeepSpeed ZeRO-3 \\
\midrule
Preference loss 
& -- 
& Sigmoid DPO, $\beta=0.1$ \\
LoRA setting 
& -- 
& rank 32; alpha 64; dropout 0.05 \\
\midrule
Workers 
& 16 preprocessing; 8 dataloader 
& 16 preprocessing; 8 dataloader \\
Distributed setup 
& 8 GPUs, single node 
& 8 GPUs, single node \\
Hardware 
& H200 GPUs 
& H200 GPUs \\
\bottomrule
\end{tabular}
\end{table}

% =============================================
\section{Preference Pair Examples}
\label{app:preference_examples}

\begin{tcolorbox}[
colback=gray!3,
colframe=gray!35,
title=\textbf{Examples of Preference Pair Construction},
fonttitle=\small,
coltitle=black,
boxrule=0.6pt,
arc=2pt,
left=5pt,
right=5pt,
top=5pt,
bottom=5pt
]
% \small
\begin{tabularx}{\linewidth}{p{0.10\linewidth} X}
\shift
&
\textbf{Chosen:} The visible fall occurs at $\sim 5.0$s, while the impact sound is heard at $\sim 3.1$s, indicating a synchronization mismatch. \\
&
\textbf{Rejected:} The sound is synchronized with the fall. \\

\specialrule{0.3pt}{4pt}{4pt}

\mute
&
\textbf{Chosen:} The audio track is silent throughout the clip; no music, speech, ambient noise, or sound effects are detected. \\
&
\textbf{Rejected:} The animated music-video scene is described as containing female vocals, a chaotic electronic rock beat, rain, thunder, TV static, and glass-shattering effects. \\

\specialrule{0.3pt}{4pt}{4pt}

\swap
&
\textbf{Chosen:} The visuals show an optics diffraction demonstration, but the audio describes how to use a centrifuge, indicating an audio-source mismatch. \\
&
\textbf{Rejected:} The narrator explains the diffraction setup as the hand shines a smartphone light through the pen tip. \\

\end{tabularx}
\end{tcolorbox}

\section{FineVideo-derived general instruction data}
\label{app:finevideo-data}

\begin{tcolorbox}[
colback=gray!2,
colframe=gray!35,
boxrule=0.6pt,
arc=2pt,
left=5pt,
right=5pt,
top=5pt,
bottom=5pt
]
\small
\vspace{4pt}

\noindent
\begin{minipage}[t]{0.49\linewidth}
\begin{tcolorbox}[
colback=descBlue,
colframe=descBlueLine,
boxrule=0.45pt,
arc=2pt,
left=5pt,
right=5pt,
top=4pt,
bottom=4pt
]
\textbf{Description}\\
Describe visible events and audible cues.
\end{tcolorbox}
\end{minipage}
\hfill
\begin{minipage}[t]{0.49\linewidth}
\begin{tcolorbox}[
colback=locGreen,
colframe=locGreenLine,
boxrule=0.45pt,
arc=2pt,
left=5pt,
right=5pt,
top=4pt,
bottom=4pt
]
\textbf{Localization}\\
Locate visual/audio events in time.
\end{tcolorbox}
\end{minipage}

\vspace{3pt}

\noindent
\begin{minipage}[t]{0.49\linewidth}
\begin{tcolorbox}[
colback=attrPurple,
colframe=attrPurpleLine,
boxrule=0.45pt,
arc=2pt,
left=5pt,
right=5pt,
top=4pt,
bottom=4pt
]
\textbf{Attribution}\\
Infer the source or material of a cue.
\end{tcolorbox}
\end{minipage}
\hfill
\begin{minipage}[t]{0.49\linewidth}
\begin{tcolorbox}[
colback=qaOrange,
colframe=qaOrangeLine,
boxrule=0.45pt,
arc=2pt,
left=5pt,
right=5pt,
top=4pt,
bottom=4pt
]
\textbf{Audio-dependent QA}\\
Answer questions that require audio evidence.
\end{tcolorbox}
\end{minipage}

\end{tcolorbox}

% ==== Appendix content ====
\section{Qualitative GPT-5.5 Outputs (Visual-Only Input)}
\label{app:gpt-qualitative}

We provide representative raw outputs of GPT-5.5 across the three tasks of our
test data---\mute, \shift, and \swap---when the model
is given visual frames only. Since the model does not have access to the
audio track, these examples illustrate how it tends to respond when forced to
reason about audio without being able to hear it.

% ---------------- Mute ----------------
\begin{tcolorbox}[gptbox,title={Task: \mute \textemdash{} What does the audio sound like?}]
\begin{tcolorbox}[videoline]
video: Bats \& Balls Fail Compilation 47.mp4
\end{tcolorbox}
\textbf{GPT-5.5 raw output:}\\
\textit{``The audio is likely from an indoor gym: basketball dribbles and
sneaker squeaks as someone runs toward the hoop, followed by the thud/clank
of a dunk attempt. Then there's a sudden loud crash as the backboard
shatters, with glass raining down onto the floor. You can hear startled
reactions---shouting, laughter, and movement---as people react and run over
afterward.''}
\end{tcolorbox}

% ---------------- Sync ----------------
\begin{tcolorbox}[gptbox,title={Task: \shift \textemdash{} Is the audio in sync with the video?}]
\begin{tcolorbox}[videoline]
video: Are You Serious! - Throwback Thursday Sept 2017 11\_delay\_2.39s.mp4
\end{tcolorbox}
\textbf{GPT-5.5 raw output:}\\
\textit{``The audio and video appear to be synchronized.\\[2pt]
The clearest sync points are the physical impact moments: the man's
jump/trip over the boards, the body hitting the ground, and the subsequent
movement/laughter reaction. The impact sound lines up with the visual fall
closely enough that I do not perceive a meaningful lead or lag. Any offset,
if present, seems very small---likely under about 0.1--0.2 seconds---and not
enough to call the tracks out of sync.''}
\end{tcolorbox}

% ---------------- Swap ----------------
\begin{tcolorbox}[gptbox,title={Task: \swap \textemdash{} Does the audio match the video content?}]
\begin{tcolorbox}[videoline]
video: Bats \& Balls Fail Compilation 47.mp4 \\
audio swapped from: FailArmy After Dark - That's Going To Cost You Ep.\,10 (103.mp4)
\end{tcolorbox}
\textbf{GPT-5.5 raw output:}\\
\textit{``yes''}
\end{tcolorbox}

\section{Evaluation Prompts}
\label{app:prompts}

This section documents the exact prompts used to elicit model responses on
each of the three tasks, together with the GPT-based judge prompts used to
parse those free-form responses into a structured prediction. For
\textsc{Mute} and \textsc{Swap}, the numbers reported in the main text
correspond to the \emph{neutral} prompt setting, in which the model is asked
an open-ended description question rather than being directly cued about the
hypothesis under test. For \textsc{Shift}, a single structured prompt is used,
which simultaneously asks the model to make an aligned/misaligned decision
and estimate the offset.

\subsection{Average Gap Calculation}
We summarize shortcut reliance by the average accuracy drop from each non-intervened control to its paired counterfactual condition:
\begin{equation}
\Delta_{\mathrm{shortcut}}
=
\frac{1}{|\mathcal{D}|}
\sum_{d \in \mathcal{D}}
\left(
\mathrm{Acc}_{\mathrm{Orig},d}
-
\mathrm{Acc}_{\mathrm{Interv},d}
\right),
\quad
\mathcal{D}=\{\mathrm{Sync}, \mathrm{Exist.}, \mathrm{Consist.}\}.
\end{equation}
Larger values indicate a larger performance collapse under counterfactual interventions. We report the gap only when all three dimensions are available. For free-form outputs, GPT-5.4 is used as an LLM judge to adjudicate predicted labels.

\subsection{Inference Prompts}

% ---- Mute ----
\begin{tcolorbox}[promptbox,title={\textsc{Mute} \textemdash{} Inference Prompt}]
\textit{``Describe the audio you hear in this video.''}
\end{tcolorbox}

% ---- Swap ----
\begin{tcolorbox}[promptbox,title={\textsc{Swap} \textemdash{} Inference Prompt}]
\textit{``Describe what you see in the video and what you hear in the audio.''}
\end{tcolorbox}

% ---- Shift ----
\begin{tcolorbox}[promptbox,title={\textsc{Shift} \textemdash{} Inference Prompt}]
\textit{``Watch this video and listen to its audio carefully. Determine
whether the audio and video tracks are synchronized. If they are not
synchronized, identify the direction of the offset (audio delayed or audio
early relative to video) and estimate the offset in seconds. Explain your
reasoning.''}
\end{tcolorbox}

\subsection{GPT Judge Prompts}

Because the inference prompts above elicit free-form text, we use GPT as a
structured-output extractor to convert each response into a discrete
prediction. The judge is prompted with the system message below and given the
model's raw response as the user message; it is required to return a single
JSON object.

% ---- Mute judge ----
\begin{tcolorbox}[judgebox,title={\textsc{Mute} \textemdash{} Judge System Prompt}]
\textit{``You are evaluating a model's response about a video whose audio has
been secretly replaced with silence. The user will give you the model's
response. Classify the model's assessment into exactly ONE of four categories
and return ONLY valid JSON (no markdown, no explanation):}\\[2pt]
\texttt{\{"prediction": "synced"|"delay"|"early"|"muted", "explanation":
"<one sentence>"\}}\\[2pt]
\textit{Rules:}
\begin{itemize}\itemsep0pt
  \item \textit{``synced'': the model claims audio and video are
        synchronized (it is hallucinating audio).}
  \item \textit{``delay'': the model claims audio is delayed relative to
        video (hallucinating audio + claiming offset).}
  \item \textit{``early'': the model claims audio is early relative to video
        (hallucinating audio + claiming offset).}
  \item \textit{``muted'': the model states there is no audio, the audio is
        silent, muted, or absent.}
  \item \textit{If the model describes specific sounds (e.g.\ ``thud'',
        ``music'', ``splash'') but does not explicitly say the video is
        silent, classify based on its sync/desync conclusion, not as
        ``muted''.}
  \item \textit{If the model is ambiguous or gives no clear answer, default
        to ``synced''.''}
\end{itemize}
\end{tcolorbox}

% ---- Swap judge ----
\begin{tcolorbox}[judgebox,title={\textsc{Swap} \textemdash{} Judge System Prompt}]
\textit{``You are evaluating a model's response about a video whose audio
has been secretly replaced with audio from a DIFFERENT unrelated video. The
user will give you the model's response. Classify the model's assessment
into exactly ONE of four categories and return ONLY valid JSON (no markdown,
no explanation):}\\[2pt]
\texttt{\{"prediction": "synced"|"delay"|"early"|"mismatched",
"explanation": "<one sentence>"\}}\\[2pt]
\textit{Rules:}
\begin{itemize}\itemsep0pt
  \item \textit{``mismatched'': the model says the audio does NOT match the
        visual content, or the audio and video are unrelated, or the sounds
        don't correspond to what's shown.}
  \item \textit{``synced'': the model claims audio and video are well
        synchronized and matching.}
  \item \textit{``delay'': the model claims audio is delayed (but still
        believes they match semantically).}
  \item \textit{``early'': the model claims audio is early (but still
        believes they match semantically).}
  \item \textit{Prefer ``mismatched'' whenever the model expresses
        confusion, doubt about the audio matching the video, or says the
        audio seems unrelated.}
  \item \textit{Default to ``synced'' if the answer is ambiguous.''}
\end{itemize}
\end{tcolorbox}

% ---- Shift judge ----
\begin{tcolorbox}[judgebox,title={\textsc{Shift} \textemdash{} Judge System Prompt}]
\textit{``You are a structured-output extractor. The user will give you a
model's free-text response about audio-video synchronization. Extract the
following fields and return ONLY valid JSON (no markdown, no explanation):}
\\[2pt]
\texttt{\{"synced": <bool>, "direction": "none"|"delay"|"early",
"offset\_sec": <float>, "t\_v": <float or null>, "t\_a": <float or null>,
"explanation": "<one sentence>"\}}\\[2pt]
\textit{Rules:}
\begin{itemize}\itemsep0pt
  \item \textit{synced: true if the model says audio and video are
        synchronized, false otherwise.}
  \item \textit{direction: ``delay'' means audio comes AFTER the visual
        event; ``early'' means audio comes BEFORE the visual event; ``none''
        if synced is true.}
  \item \textit{offset\_sec: estimated time gap in seconds. 0.0 if synced.}
  \item \textit{t\_v: the timestamp (in seconds) the model attributes to
        the VISUAL event. null if not mentioned.}
  \item \textit{t\_a: the timestamp (in seconds) the model attributes to
        the AUDIO event. null if not mentioned.}
  \item \textit{If you cannot determine a field, use the default (true /
        ``none'' / 0.0 / null / ``'').''}
\end{itemize}
\end{tcolorbox}

%================================
\section{Failure-mode definitions}
\label{app:failure-modes}

This appendix provides the detailed definition and measurement protocol for
each of the eight failure modes reported in
\Cref{fig:failure_heatmap}. All rates lie in $[0,1]$; higher values
indicate more frequent failures. Free-form (neutral-prompt) responses are
classified by an independent OpenAI GPT-5.4 judge so that the mute/swap modes
do not depend on a model self-reporting its own confusion.

\paragraph{A. Audio Hallucination.}
Errors in which the model invents or accepts audio content that is
incompatible with the input.

\begin{description}
  \item[Mute Hallucination.] On videos whose audio track has been replaced
    with silence, the prompt is \textit{``Describe the audio you hear in this
    video.''} The judge classifies each response into
    \texttt{muted} / \texttt{audio\_described} / \texttt{visual\_only}, and
    \emph{Mute Hallucination} is the \texttt{audio\_described} rate: the
    fraction of responses in which the model produces any concrete description
    of audio content (speech, music, ambient noise, impacts) instead of
    reporting silence.

  \item[Swap False-Match.] On videos whose audio track has been replaced with
    the soundtrack of an unrelated video, the prompt is \textit{``Describe
    what you see in the video and what you hear in the audio.''}
    \emph{Swap False-Match} is the fraction of responses in which the judge
    concludes the model treated the (mismatched) audio as a plausible natural
    match for the visuals.
\end{description}

\paragraph{B.\ Audio Denial.}
Symmetric errors on naturally paired (un-intervened) videos.

\begin{description}
  \item[False Silence.] On videos with their original audio, the same neutral
    mute prompt is used. \emph{False Silence} is the fraction of responses in
    which the model claims silence or ``no audible content'' despite real
    audio being present.

  \item[Swap False-Mismatch.] On videos with their original audio, the same
    neutral swap prompt is used. \emph{Swap False-Mismatch} is the fraction
    of responses in which the model spuriously claims an audio--visual
    mismatch on a naturally synchronized pair.
\end{description}

\paragraph{C.\ Question Avoidance.}

\begin{description}
  \item[Audio Dodge.] Fraction of responses to the neutral mute prompt in
    which the model produces a visual-only description and never engages with
    the audio question (neither describes any sound nor claims silence). We
    report the mean of this rate across the intervention (silenced) and
    control (real audio) conditions because non-engagement is a property of
    the model, not of whether audio is real.
\end{description}

\paragraph{D.\ Temporal Failures (sync task).}
The sync benchmark contains synced originals together with two intervention
variants per video: an audio-delayed copy (\textsc{delay}) and an
audio-advanced copy (\textsc{early}), both with offsets of approximately
$\pm 2$\,s. Each model's free-form response is parsed by the judge into a
boolean \texttt{pred\_synced} and a categorical
$\texttt{pred\_direction}\in\{\textsc{delay},\textsc{early},\textsc{none}\}$.

\begin{description}
  \item[Offset Blindness.] Fraction of desync samples (delay or early) for
    which the model judges the clip to be synced. This isolates pure failure
    to perceive temporal misalignment.

  \item[Direction Confusion.] Among the desync samples on which the model
    correctly judges the clip to be non-synced, the fraction for which it
    picks the \emph{wrong} direction (calls a delay an early or vice versa).
    This isolates direction-of-offset perception from offset detection
    itself.

  \item[False Sync Alarm.] Fraction of synced original samples for which the
    model claims the clip is desynced. Symmetric counterpart to Offset
    Blindness: false alarms on naturally aligned audio.
\end{description}

For Gemini-3.1-pro, per-row sync predictions were not retained, so its
Offset Blindness, Direction Confusion, and False Sync Alarm are derived from
the aggregate \texttt{sync\_desync\_accuracy},
\texttt{direction\_accuracy\_on\_desync}, and per-category accuracies in the
saved \texttt{metrics.json}, which are mathematically equivalent to the
per-row counts.

% ================================================
\section{Limitations.}
\label{app:limit}
Our training recipe is currently evaluated on a limited set of base models, so its effectiveness across broader omni-modal model families remains to be further studied. In addition, our recipe experiments primarily validate the effect of applying DPO after SFT for improving temporal synchronization. We have not yet conducted a complete training study for the Mute and Swap settings, which probe audio existence and cross-modal consistency. Extending the recipe to these intervention types is an important direction for future work.

\section{Ethics and Broader Impacts}
\label{app:ethics-impact}

\subsection{Ethics.}
Our research follows the NeurIPS Code of Ethics. The study is designed as a diagnostic and alignment analysis of audio-visual grounding in multimodal models, and does not involve human-subject experiments, crowdsourcing, or the collection of personally identifiable information. The video data used in our experiments comes from public or properly licensed sources, and we use the data only for model evaluation and training under controlled audio-visual interventions. Our analysis does not perform face recognition, identity inference, biometric classification, or any other person-level profiling.

\subsection{Broader Impacts}
\paragraph{Positive impacts.}
This work aims to improve the reliability of video-capable multimodal and omni-modal models by revealing when they rely on visual-semantic shortcuts rather than genuine audio-visual verification. By introducing controlled Mute, Swap, and Shift interventions, our evaluation can help researchers diagnose pseudo-alignment and develop models that more faithfully check whether audio is present, synchronized, and consistent with the visual scene. Such improvements may benefit downstream applications where audio-visual grounding is important, including assistive technologies, video understanding, human-computer interaction, and safety-critical multimodal monitoring.

\paragraph{Potential risks and mitigations.}
The main risk is that intervention-based diagnostics could be used to construct adversarial examples or to optimize models specifically for benchmark performance rather than robust real-world grounding. In addition, improved audio-visual verification does not eliminate all hallucination risks, and deployed systems may still fail under out-of-distribution sounds, noisy environments, edited videos, or subtle cross-modal inconsistencies. To mitigate these risks, we frame our benchmark as a diagnostic tool rather than a deployment guarantee, report limitations of the evaluated settings, and encourage evaluating models under diverse interventions instead of relying on aggregate accuracy alone. If releasing data or code, we will document intended use, licenses, and limitations, and avoid releasing sensitive or personally identifying content.

\section{New Assets}
\label{app:assets}

We introduce intervention-based evaluation assets for probing audio-visual grounding under three controlled settings: Mute, Swap, and Shift. These assets are intended for diagnostic evaluation, testing whether multimodal models verify the audio stream rather than relying on visual-semantic shortcuts.

For each selected video, we construct intervention variants by modifying only the audio stream. Mute removes the audio track to test audio-existence verification. Swap replaces the original audio with mismatched audio to test cross-modal consistency. Shift temporally displaces the audio to test synchronization and offset reasoning. Each example is paired with its intervention type, evaluation condition, and target label.

The assets are verified by members of the research team to ensure that the intervention is valid and the label is unambiguous. We reject examples with unclear visual events, corrupted or inaudible audio, failed interventions, or ambiguous labels. The assets are used only for model evaluation and alignment research, and are not intended as a guarantee of real-world robustness. Their limitations include restricted intervention coverage, possible residual annotation noise, and limited coverage of all real-world audio-visual failure modes.